\documentclass[runningheads]{llncs}
\usepackage{graphicx}
\usepackage{comment}
\usepackage{color}
\usepackage{epsfig}
\usepackage{graphicx}
\usepackage{amsmath}
\usepackage{amssymb}
\usepackage[ruled,vlined]{algorithm2e}

\usepackage[utf8]{inputenc}
\usepackage{float}
\usepackage[export]{adjustbox}
\usepackage{subcaption}
\usepackage{graphics}
\usepackage{makecell}
\usepackage{multirow}
\usepackage{flushend}
\usepackage{xcolor,soul,colortbl}
\usepackage{sidecap}
\usepackage{wrapfig}
\usepackage{booktabs}       
\usepackage{footnote}
\usepackage{authblk}
\makesavenoteenv{tabular}
\makesavenoteenv{table}
\DeclareMathAlphabet{\mathbbold}{U}{bbold}{m}{n}

\usepackage[pagebackref=true,breaklinks=true,letterpaper=true,bookmarks=false]{hyperref}

\begin{document}
\pagestyle{headings}
\mainmatter
\def\ECCVSubNumber{XXXX}  

\title{Toward Unsupervised, Multi-Object Discovery\\in Large-Scale Image Collections}
\titlerunning{Toward unsupervised, multi-object discovery\\in large-scale image collections}
\author{Huy V. Vo\inst{1,2,3} \and Patrick P{\'e}rez \inst{3} \and Jean Ponce \inst{1,2}}
\authorrunning{H. V. Vo {\em et al.}}
\institute{INRIA, Paris, France \and D{\'e}partement  d'informatique de l'ENS, ENS, CNRS, PSL University, Paris, France \and Valeo.ai}
\maketitle

\begin{abstract}
This paper addresses the problem of discovering the objects present in a collection of images without any supervision. We build on the optimization approach of Vo {\em et al.} [34] with several key novelties: (1) We propose a novel saliency-based region proposal algorithm that achieves significantly higher overlap with ground-truth objects than other competitive methods. This procedure leverages off-the-shelf CNN features trained on classification tasks without any bounding box information, but is otherwise unsupervised. (2) We exploit the inherent hierarchical structure of proposals as an effective regularizer for the approach to object  discovery of [34], boosting its performance to significantly improve over the state of the art on several standard benchmarks. (3) We adopt a two-stage strategy to select promising proposals using small random sets of images before using the whole image collection to discover the objects it depicts, allowing us to tackle, for the first time (to the best of our knowledge), the discovery of multiple objects in each one of the pictures making up datasets with up to 20,000 images, an over five-fold increase compared to existing methods, and a first step toward true large-scale unsupervised image interpretation. 
\end{abstract}

\section{Introduction}
Object discovery, that is finding the location of salient objects in images without using any source of supervision, is a fundamental scientific problem in computer vision. It is also potentially an important practical one, since any effective solution would serve as a reliable free source of supervision for other tasks such as object categorization, object detection and the like. While many of these tasks can be tackled using massive amounts of annotated data, the manual annotation process is complex and expensive at large scales. Combining the discovery results with a limited amount of annotated data in a semi-supervised setting is a promising alternative to current data-hungry supervised approaches~\cite{Wei2019ddtplus}.

Vo {\em et al.}~\cite{Vo2019UnsupOptim} posit that image collections possess an implicit graph structure.
The pictures themselves are the nodes, and an edge links two images when they share similar visual content. They propose the object and structure discovery framework (OSD) to localize objects and find the graph structure simultaneously by solving an optimization problem. Though demonstrating promising results,~\cite{Vo2019UnsupOptim} has several shortcomings, e.g., the use of supervised region proposals, the limitation in addressing large image collections (See Section~\ref{sec:related_work}). Our work is built on OSD, aims to alleviate its limitations and improves it to effectively discover multiple objects in large image collections. Our contributions are:

\indent{$\bullet$} We propose a simple but effective method for generating region proposals directly from CNN features (themselves trained beforehand on some auxiliary task~\cite{Symonian2014verydeep} {\em without} bounding boxes) in an unsupervised way (Section~\ref{sec:unsup_region_proposals}). Our algorithm gives on average half the number of region proposals per image compared to selective search~\cite{uijlings2013selective}, edgeboxes~\cite{zitnick2014edge} or randomized Prim~\cite{Manen2013prim}, yet significantly outperforms these off-the-shelf region proposals in object discovery (Table~\ref{table:separate_compare_proposals}). 
\indent{$\bullet$} Leveraging the intrinsic structure of region proposals generated by our method allows us to add an additional constraint into the OSD formulation that acts as a regularizer on its behavior (Section~\ref{sec:rOSD}). This new formulation (rOSD) significantly outperforms the original algorithm and allows us to effectively perform multi-object discovery, a setting never studied before (to the best of our knowledge) in the literature.

\indent{$\bullet$} We propose a two-stage algorithm to make rOSD applicable to large image collections (Section~\ref{sec:large_scale_osd}). In the first stage, rOSD is used to choose a small set of good region proposals for each image. In the second stage, these proposals and the full image collection are fed to rOSD to find the objects and the image graph structure.

\indent{$\bullet$} We demonstrate that our approach yields significant improvements over the state of the art in object discovery (Tables~\ref{table:single_separate} and~\ref{table:single_mixed}). We also run our two-stage algorithm on a new and much larger dataset with 20,000 images and show that it significantly outperforms plain OSD in this setting (Table~\ref{table:large_scale_performance}).

The only supervisory signal used in our setting are the image labels used to train CNN features in an auxiliary classification task (see~\cite{Li2016mimick,Wei2019ddtplus} for similar approaches in the related colocalization domain). We use CNN features trained on ImageNet classification~\cite{Symonian2014verydeep}, {\em without} any bounding box information. Our region proposal and object discovery algorithms are otherwise fully unsupervised.

\section{Related Work}
\label{sec:related_work}
Region proposals have been used in object detection/discovery to serve as object priors and reduce the search space. In most cases, they are found either by a bottom-up approach in which low-level cues are aggregated to rank a large set of boxes obtained with sliding window approaches~\cite{alexe2012measuring,uijlings2013selective,zitnick2014edge} and return the top windows as proposals, or by training a model to classify them (as in randomized Prim~\cite{Manen2013prim}, see also~\cite{ren15fasterrcnn}), with \textit{ bounding box supervision}. Edgeboxes~\cite{zitnick2014edge} and selective search~\cite{uijlings2013selective} are popular off-the-shelf algorithms that are used to generate region proposals in object detection~\cite{girshickICCV15fastrcnn,girshick2014rcnn}, weakly supervised object detection~\cite{Cinbis2015weakly,Tang2018TPAMI_pcl_wosd} or image colocalization~\cite{Li2016mimick}. Note, however, that the features used to generate proposals in these algorithms and those representing them in the downstream tasks are generally different in nature: Typically, region proposals are generated from low-level features such as color and texture~\cite{uijlings2013selective} or edge density~\cite{zitnick2014edge}, but CNN features are used to represent them in downstream tasks. However, the Region Proposal Network in Faster-RCNN~\cite{ren15fasterrcnn} shows that proposals generated directly from the features used in the object detection task itself give a great boost in performance. In the object discovery setting, we therefore propose a novel approach for generating region proposals in an unsupervised way from CNN features trained on an auxiliary classification task without bounding box information. Features from CNNs trained on large-scale image classification have also been used to localize object in the weakly supervised setting. Zhou {\em et al.}~\cite{Zhou2016cvpr} and Selvaraju {\em et al.}~\cite{Selvaraju2017ICCV} fine-tune a pre-trained CNN to classify images and construct class activation maps, as weighted sums of convolutional feature maps or their gradient with respect to the classification loss, for localizing objects in these images. Tang {\em et al.}~\cite{Tang2018weakRPN} generate region proposals to perform weakly supervised object detection on a set of labelled images by training a proposal network using the images' labels as supervision. Contrary to these works, we generate region proposals using only pre-trained CNN features without fine-tuning the feature extractor. Moreover, our region proposals come with a nice intrinsic structure which can be exploited to improve object discovery performance. 

Early work on object discovery~\cite{faktor2012clustering,grauman2006unsupervised,kim2009unsupervised,Russell06,ICCV/SivicREZF05} focused on a restricted setting where images are from only a few distinctive object classes.  Cho {\em et al.}~\cite{CKSP15} propose an approach for object and structure discovery by combining a part-based matching technique and an iterative \textit{match-then-localize} algorithm, using off-the-shelf region proposals as primitives for matching. Vo {\em et al.}~\cite{Vo2019UnsupOptim} reformulate~\cite{CKSP15} in an optimization framework and obtain significantly better performance. Image colocalization can be seen as a narrow setting of object discovery where all images in the collection contain objects from the same class. Observing that supervised object detectors often assign high scores to only a small number of region proposals, Li {\em et al.}~\cite{Li2016mimick} propose to mimic this behavior by training a classifier to minimize the entropy of the scores it gives to region proposals. Wei {\em et al.}~\cite{Wei2019ddtplus} localize objects by clustering pixels with high activations in feature maps from CNNs pre-trained in ImageNet. All of the above works, however, focus on discovering only the main object in the images and target small-to-medium-scale datasets. Our approach is based on a modified version of the OSD formulation of Vo {\em et al.}~\cite{Vo2019UnsupOptim} and pre-trained CNN features for object discovery, offers an effective and efficient solution to discover multiple objects in images in large-scale datasets. The recent work of Hsu {\em et al.}~\cite{HsuCVPR19DeepCO3} for instance co-segmentation can also be adapted for localizing multiple objects in images. However, it requires input images to contain an object of a single dominant class while images may instead contain several objects from different categories in our setting.

\noindent{\textit{Object and structure discovery (OSD)~\cite{Vo2019UnsupOptim}.}} Since our work is built on~\cite{Vo2019UnsupOptim}, we give a short recap of this work in this section. Given a collection of $n$ images, possibly containing objects from different categories, each equipped with $p$ region proposals (which can be obtained using selective search~\cite{uijlings2013selective}, edgeboxes~\cite{zitnick2014edge}, randomized Prim~\cite{Manen2013prim}, etc.) and a set of potential neighbors, the unsupervised object and structure discovery problem (OSD) is formalized in~\cite{Vo2019UnsupOptim} as follows: Let us define the variable $e$ as an element of $\{0,1\}^{n\times n}$ with a zero diagonal, such that $e_{ij} = 1$ when images $i$ and $j$
are linked by a (directional) edge, and $e_{ij} = 0$ otherwise, and the variable $x$ as an element of $\{0,1\}^{n\times p}$, with $x^k_i = 1$ when region proposal number $k$ corresponds to visual content shared with neighbors of image $i$ in the graph. This leads to the following optimization problem:
\begin{equation}
\!\!\!\!\!\!\underset{x,e}{\mathrm{max}}\   
S(x,e)=\!\!\sum_{i=1}^n \!\sum\limits_{j \in N(i)}\!\!\! e_{ij} x_i^T S_{ij} x_j, \ 
\text{s.t.}\sum\limits_{k=1}^p x_i^k\le \nu \text{ and } \sum\limits_{j\neq i} e_{ij}\le\tau \ \forall i,
\label{eq:main}
\end{equation}
where $N(i)$ is the set of potential neighbors of image $i$, $S_{ij}$ is a $p\times p$ matrix whose entry $S_{ij}^{kl}$
measures the similarity between regions $k$ and $l$ of images $i$ and
$j$, and $\nu$ and $\tau$ are predefined constants corresponding
respectively to the maximum number of objects present in an image and
to the maximum number of neighbors an image may have.
This is however a hard combinatorial optimization problem. As shown in~\cite{Vo2019UnsupOptim}, an approximate solution can be found by (a) a dual gradient ascent algorithm for a continuous relaxation of Eq.\,(\ref{eq:main}) with exact updates obtained by maximizing a supermodular cubic pseudo-Boolean function~\cite{Bach13,NeOz09}, (b) a simple greedy scheme, or (c) a combination thereof. Since solving the continuous relaxation of Eq.\,(\ref{eq:main}) is computationally expensive and may be less effective for large datasets~\cite{Vo2019UnsupOptim}, we only consider the version (b) of OSD in our analysis.

 OSD has some limitations: (1) Although the algorithm itself is fully unsupervised, it gives by far its best results with region proposals from randomized Prim~\cite{Manen2013prim}, a region proposal algorithm trained with bounding box supervision. (2) Vo {\em et al.} use whitened HOG (WHO)~\cite{hariharan2012discriminative} to represent region proposals in their implementation although CNN features work better on the similar image colocalization problem~\cite{Li2016mimick,Wei2019ddtplus}. In our experiments, naively switching to CNN features does not give consistent improvement on common benchmarks. OSD with CNN features gives a CorLoc of $82.9$, $71.5$ and $42.8$ compared to $87.1$, $71.2$ and $39.5$ given by OSD with WHO, respectively on OD, VOC\_6x2 and VOC\_all data sets respectively.  (3) Finally, due to its high memory cost, the algorithm cannot be applied to large datasets without compromising its final performance. In the next section, we describe our approach to addressing these limitations, as well as extending OSD to solve multi-object discovery.

\section{Proposed Approach}
\subsection{Region Proposals from CNN Features}
\label{sec:unsup_region_proposals}
\begin{figure*}[tb]
\captionsetup[subfigure]{labelformat=empty}
	\begin{center}
	    \begin{subfigure}{0.135\textwidth}
	        \includegraphics[width=\linewidth]{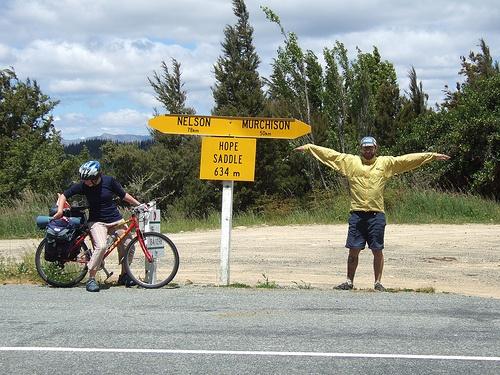}
		\end{subfigure}
		\begin{subfigure}{0.135\textwidth}
	        \includegraphics[width=\linewidth]{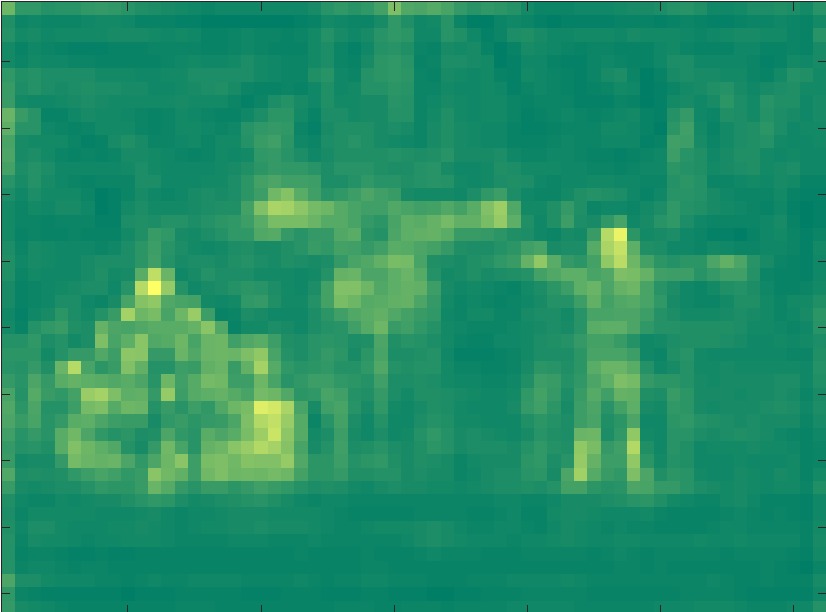}
		\end{subfigure}
		\begin{subfigure}{0.135\textwidth}
			\includegraphics[width=\linewidth]{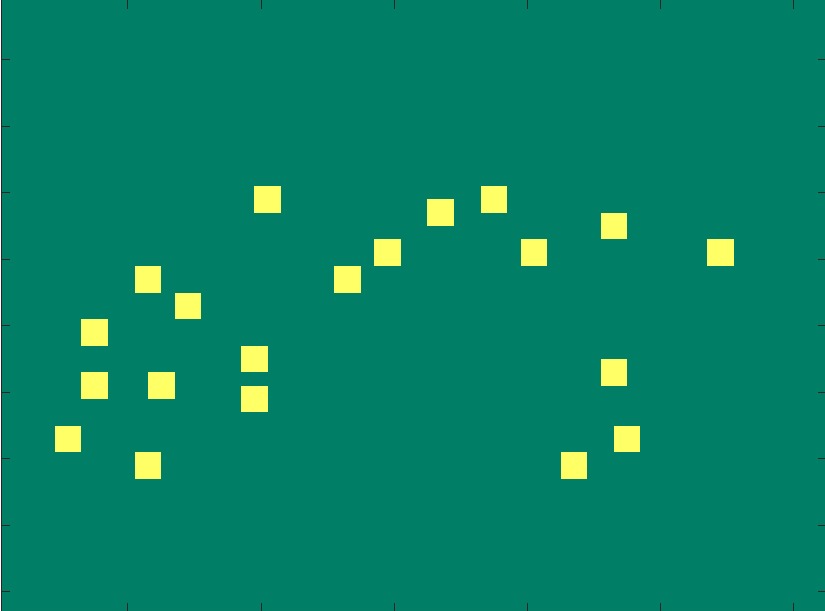}
		\end{subfigure}
		\begin{subfigure}{0.135\textwidth}
			\includegraphics[width=\linewidth]{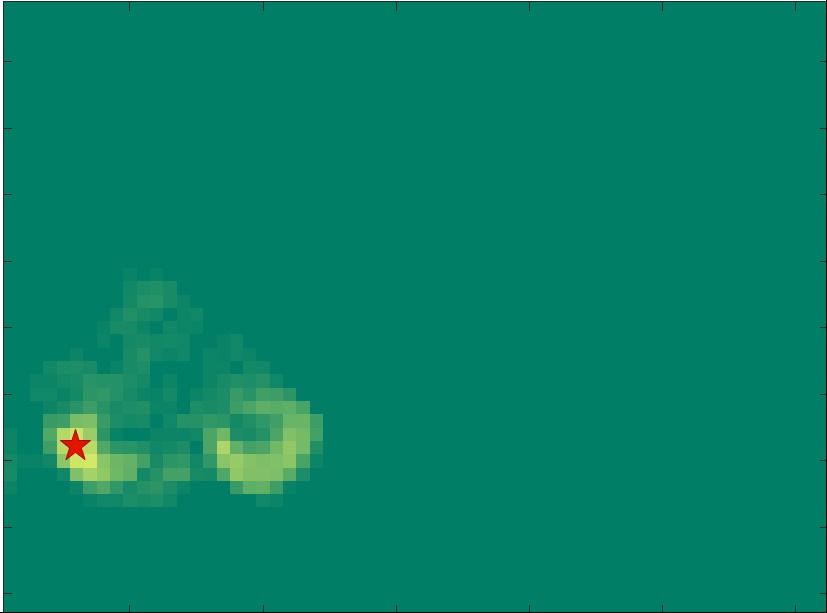}
		\end{subfigure}
		\begin{subfigure}{0.135\textwidth}
			\includegraphics[width=\linewidth]{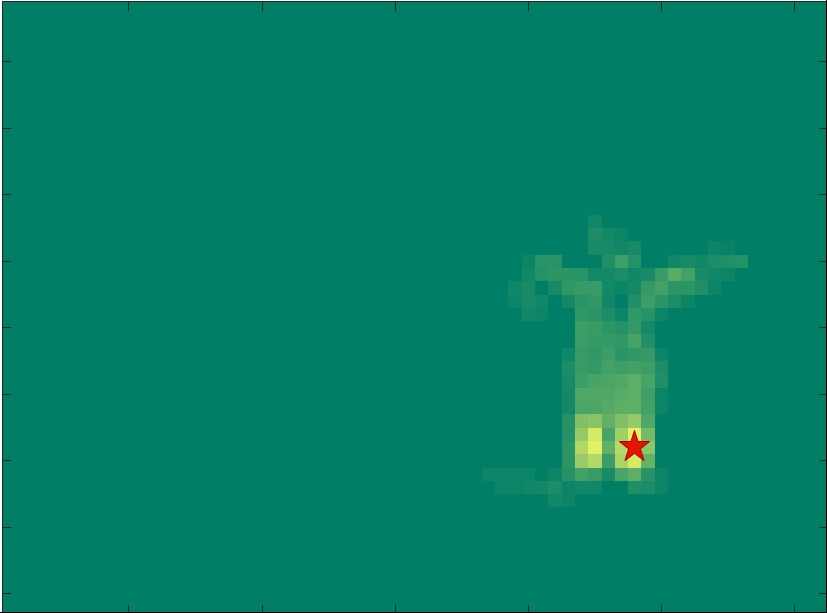}
		\end{subfigure}
		\begin{subfigure}{0.135\textwidth}
			\includegraphics[width=\linewidth]{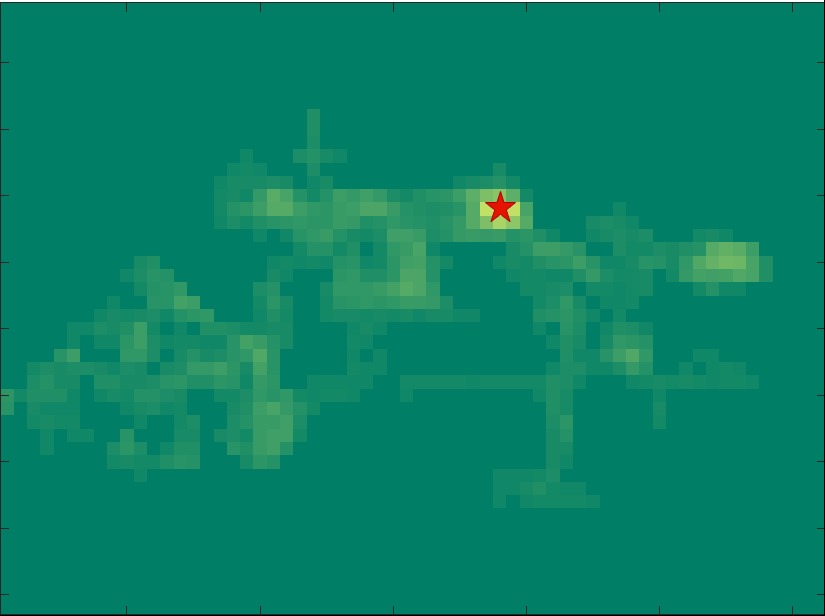}
		\end{subfigure}
		
		\vskip 0.2em
		\begin{subfigure}{0.135\textwidth}
			\includegraphics[width=\linewidth]{images/bicycle_right_1_saliency_local_16.jpg}
		\end{subfigure}
		\begin{subfigure}{0.135\textwidth}
			\includegraphics[width=\linewidth]{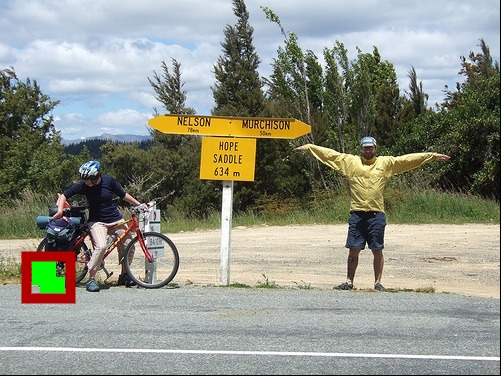}
		\end{subfigure}
		\begin{subfigure}{0.135\textwidth}
			\includegraphics[width=\linewidth]{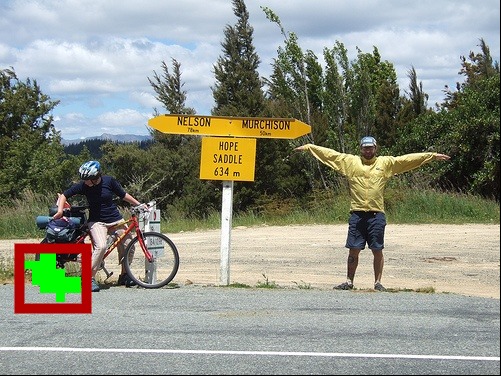}
		\end{subfigure}
		\begin{subfigure}{0.135\textwidth}
			\includegraphics[width=\linewidth]{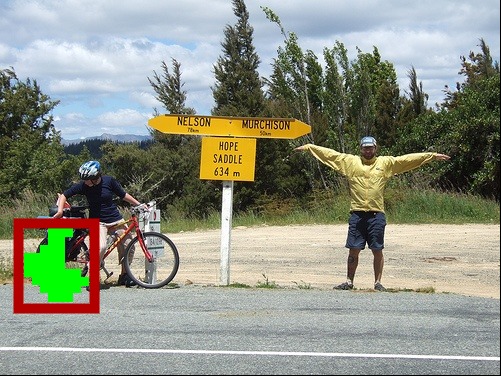}
		\end{subfigure}
		\begin{subfigure}{0.135\textwidth}
			\includegraphics[width=\linewidth]{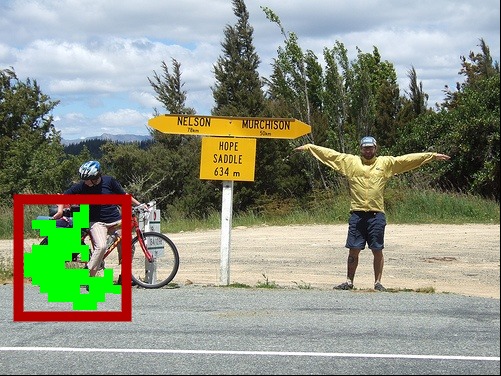}
		\end{subfigure}
		\begin{subfigure}{0.135\textwidth}
			\includegraphics[width=\linewidth]{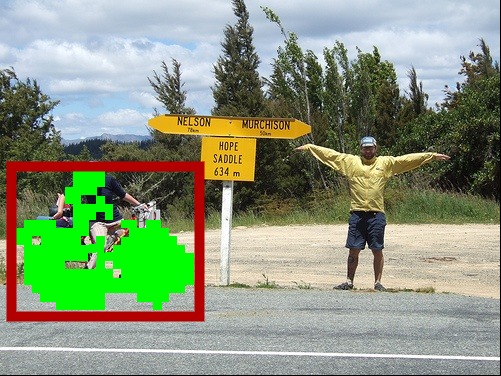}
		\end{subfigure}
		
		\begin{subfigure}{0.135\textwidth}
			\includegraphics[width=\linewidth]{images/bicycle_right_1_saliency_local_3.jpg}
		\end{subfigure}
		\begin{subfigure}{0.135\textwidth}
			\includegraphics[width=\linewidth]{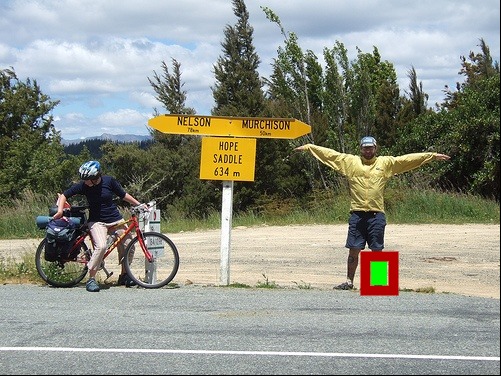}
		\end{subfigure}
		\begin{subfigure}{0.135\textwidth}
			\includegraphics[width=\linewidth]{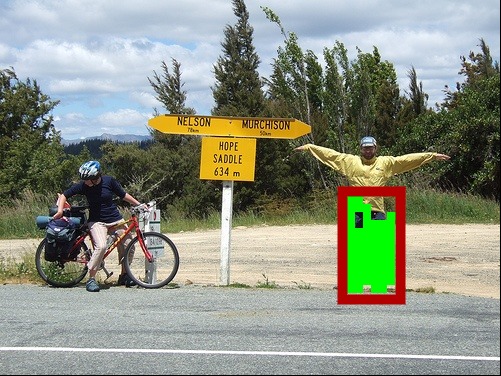}
		\end{subfigure}
		\begin{subfigure}{0.135\textwidth}
			\includegraphics[width=\linewidth]{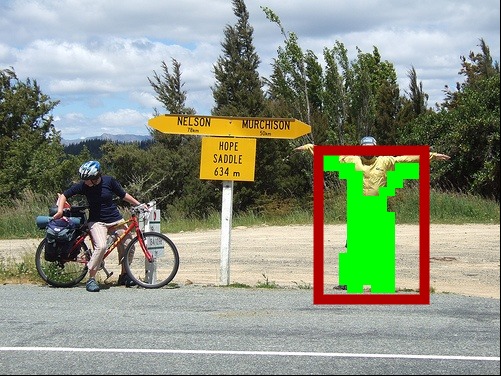}
		\end{subfigure}
		\begin{subfigure}{0.135\textwidth}
			\includegraphics[width=\linewidth]{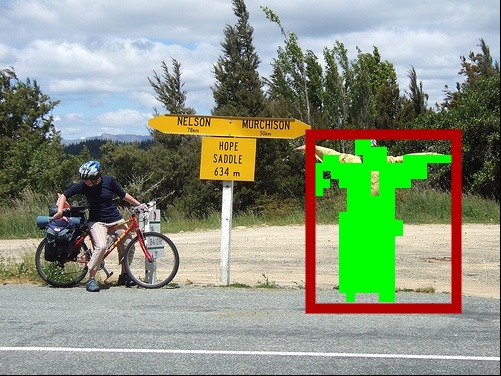}
		\end{subfigure}
		\begin{subfigure}{0.135\textwidth}
			\includegraphics[width=\linewidth]{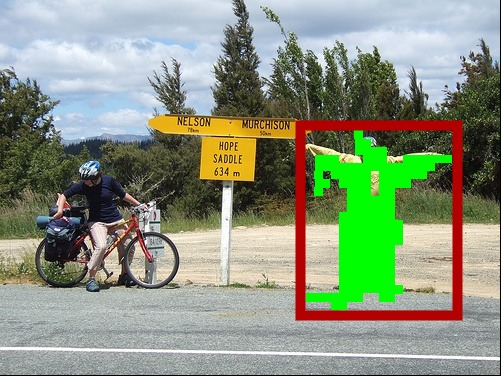}
		\end{subfigure}
		
		\begin{subfigure}{0.135\textwidth}
			\includegraphics[width=\linewidth]{images/bicycle_right_1_saliency_local_6.jpg}
		\end{subfigure}
		\begin{subfigure}{0.135\textwidth}
			\includegraphics[width=\linewidth]{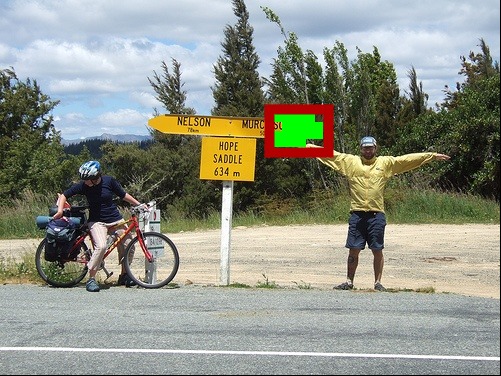}
		\end{subfigure}
		\begin{subfigure}{0.135\textwidth}
			\includegraphics[width=\linewidth]{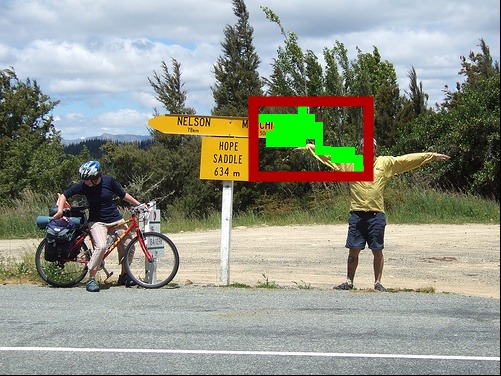}
		\end{subfigure}
		\begin{subfigure}{0.135\textwidth}
			\includegraphics[width=\linewidth]{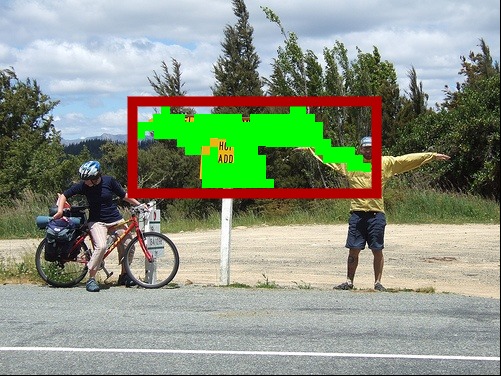}
		\end{subfigure}
		\begin{subfigure}{0.135\textwidth}
			\includegraphics[width=\linewidth]{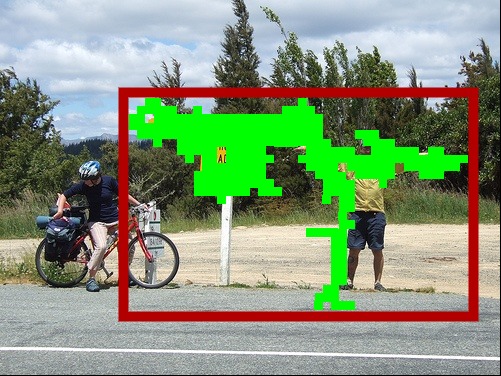}
		\end{subfigure}
		\begin{subfigure}{0.135\textwidth}
			\includegraphics[width=\linewidth]{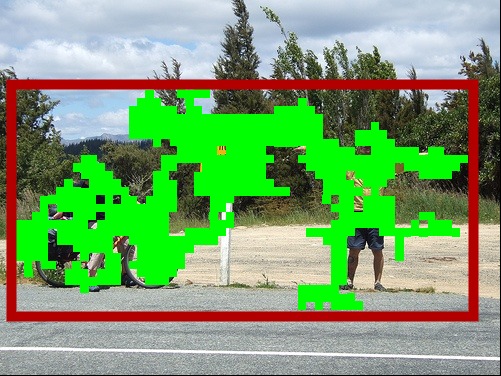}
		\end{subfigure}

	\end{center}
	\vspace{-0.4cm}
	\caption{\small Illustration of the unsupervised region proposal generation process. The top row shows the original image, the global saliency map $s_g$, local maxima of $s_g$ and three local saliency maps $s_y$ from three local maxima (marked by red stars). The next three rows illustrate the proposal generation process on the local saliency maps: From left to right, we show in green the connected component formed by pixels with saliency above decreasing thresholds and, in red, the corresponding region proposals. 
	}
	\label{fig:unsup_proposals}
	\vspace{-0.6cm}
\end{figure*}

We address the limitation of using off-the-shelf region proposals of~\cite{Vo2019UnsupOptim} with insights gained from the remarkably effective method for image colocalization proposed by Wei {\em et al.}~\cite{Wei2019ddtplus}: CNN features pre-trained for an auxiliary task, such as ImageNet classification, give a strong, {\em   category-independent} signal for unsupervised tasks. In retrospect, this insight is not particularly surprising, and it is implicit in several successful approaches to image retrieval~\cite{zhang2015self_pace} or co-saliency detection~\cite{Babenko2014neuralcode,Babenko2015sumaggregate,Hsu2018cosaliency,Wei2017scda}. Wei {\em et  al.}~\cite{Wei2019ddtplus} use it to great effect in the image colocalization task. Feeding an image to a pre-trained convolutional neural network yields a set of feature maps represented as a 3D tensor (e.g., a convolutional layer of VGG16~\cite{Symonian2014verydeep} or ResNet~\cite{He16}). Wei {\em et al.}~\cite{Wei2019ddtplus} observe that the ``image'' obtained by simply adding the feature maps gives hints to the locations of the objects it contains, and identify objects by clustering pixels with high activation. Similar but different from them, we observe that local maxima in the above ``images'' correspond to salient parts of objects in the original image and propose to exploit this observation for generating region proposals directly from CNN features. As we do not make use of any annotated bounding boxes, our region proposal itself is indeed unsupervised. Our method consists of the following steps. First, we feed the image to a pre-trained convolutional neural network to obtain a 3D tensor of size $(H\times W\times D)$, noted $F$. Adding elements of the tensor along its depth dimension yields a $(H\times W)$ 2D saliency map, noted as $s_g$ (\textit{global} saliency map), showing salient locations in the image with each location in $s_g$ being represented by the corresponding $D$-dimensional feature vector from $F$. Next, we find robust local maxima in the previous saliency map using \textit{persistence}, a measure used in topological data analysis~\cite{Chazal2013persistence,Edelsbrunner2009introtopo,Edelsbrunner2002topo,Oudot2015persistence,Zomorodian2005compute} to find critical points of a function (see Section~\ref{sec:implementation} for details). We find regions around each local maximum $y$ using a \textit{local} saliency map $s_y$ of the same size as the global one. The value at any location in $s_y$ is the dot product between normalized feature vectors at that location and the local maximum. By construction, the local saliency map highlights locations that are likely to belong to the same object as the corresponding local maximum. Finally, for each local saliency map, we discard all locations with scores below some threshold and the bounding box around the connected component containing the corresponding local maximum is returned as a region proposal. By varying the threshold, we can obtain tens of region proposals per local saliency map. An example illustrating the whole process is shown in Fig.~\ref{fig:unsup_proposals}.

\subsection{Regularized OSD}
\label{sec:rOSD}
Due to the greedy nature of OSD~\cite{Vo2019UnsupOptim}, its block-coordinate ascent iterations are prone to bad local maxima. Vo {\em et al.}~\cite{Vo2019UnsupOptim} attempt to resolve this problem by using a larger value of $\nu$ in the optimization than the actual number of objects they intend to retrieve (which is one in their case) to diversify the set of retained regions in each iteration. The final region in each image is then chosen amongst its retained regions in a post processing step by ranking these using a new score solely based on their similarity to the retained regions in the image's neighbors. Increasing $\nu$ in fact gives limited help in diversifying the set of retained regions. Since there is redundancy in object proposals with many highly overlapping regions, the $\nu$ retained regions are often nearly identical (see supplementary document for a visual illustration). This phenomenon also prevents OSD from retrieving multiple objects in images. One can use the ranking in OSD's post processing step with non-maximum suppression to return more than one region from $\nu$ retained regions but since $\nu$ regions are often highly overlapping, this fails to localize multiple objects. 

By construction, proposals produced by our approach also contain many highly overlapping regions, especially those generated from the same local maximum in the saliency map. However, they come with a nice intrinsic structure: Proposals in an image can be partitioned into groups labelled by the local maximum from which they are generated. Naturally, it makes sense to impose that at most one region in a group is retained in OSD since they are supposed to correspond to the same object. This additional constraint also conveniently helps to diversify the set of proposals returned by the block-coordinate ascent procedure by avoiding to retain highly overlapping regions. Concretely, let $G_{ig}$ be the set of region proposals in image $i$ generated from the $g$-th local maximum in its global saliency map $s_g$, with $1 \le g \leq L_i$ where $L_i$ is the number of local maxima in $s_g$, we propose to add the constraints $\sum_{k \in G_{ig}} x_i^k \le 1\ \forall i,g$ to Eq.\,(\ref{eq:main}).
We coin the new formulation regularized OSD (rOSD). Similar to OSD, a solution to rOSD can be obtained by a greedy block-coordinate ascent algorithm whose iterations are illustrated 
in the supplementary document. We will demonstrate the effectiveness of rOSD compared to OSD and the state of the art in Section~\ref{sec:experiments}.

\subsection{Large-Scale Object Discovery}
\label{sec:large_scale_osd}
The optimization algorithm of Vo {\em et al.}~\cite{Vo2019UnsupOptim} requires loading all score matrices $S_{ij}$ into the memory (they can also be computed on-the-fly but at an unacceptable computational cost). The corresponding memory cost is $M = (\sum_{i=1}^n |N(i)|) \times K$, decided by two main factors: The number of image pairs considered $\sum_{i=1}^n |N(i)|$ and the number of positive entries $K$ in matrices $S_{ij}$. To reduce the cost on larger datasets, Vo {\em et al.}~\cite{Vo2019UnsupOptim} pre-filter the neighborhood of each image ($|N(i)| \leq 100$ for classes with more than 1000 images) and limit $K$ to $1000$. This value of $K$ is approximately the average number of proposals in each image, and it is intentionally chosen to make sure that $S_{ij}$ is not too sparse in the sense that approximately every proposal in image $i$ should have a positive match with some proposal in image $j$. Further reducing the number of positive entries in score matrices is likely to hurt the performance (Table~\ref{table:large_scale_performance}) while a number of 100 potential neighbors is already small and can not be significantly lowered. Effectively scaling up OSD\footnote{Since the analysis in this section applies to both OSD and rOSD, we refer to both as OSD for ease of notation.} therefore requires lowering considerably the number of proposals it uses. To this end, we propose two different interpretations of the image graph and exploit both to scale up OSD.

\noindent{\textit{Two different interpretations of the image graph.}}
The {\em image graph} $G=(x,e)$ obtained by solving
Eq.\,(\ref{eq:main}) can be interpreted as capturing the ``true'' structure of the input image collection. In this case, $\nu$ is typically small (say, 1 to 5) and the discovered ``objects'' correspond to maximal cliques of $G$, with instances given by active regions ($x_i^k~=~1$) associated with nodes in the clique. But it can also be interpreted as a {\em proxy} for that structure. In this case, we typically take $\nu$ larger (say, 50). The active regions found for each node $x_i$ of $G$ are interpreted as the most promising regions in the corresponding image and the active edges $e_{ij}$ link it to other images supporting that choice. We dub this variant {\em proxy} OSD.

For small image collections, it makes sense to run OSD only. For large ones, we propose instead to split the data into random groups with fixed size, run proxy OSD on each group to select the most promising region proposals in the corresponding images, then run OSD using these proposals. Using this two-stage algorithm, we reduce significantly the number of image pairs in each run of the first stage, thus permitting the use of denser score matrices in these runs. In the second stage, since only a very small number of region proposals are considered in each image, we need to keep only a few positive entries in each score matrix and are able to run OSD on the entire image collection. Our approach for large-scale object discovery is summarized the supplementary material.

\section{Experiments}
\label{sec:experiments}

\subsection{Datasets and Metrics}
\label{sec:datasets_and_metrics}
Similar to previous works on object discovery~\cite{CKSP15,Vo2019UnsupOptim} and image colocalization \cite{Li2016mimick,Wei2019ddtplus}, we evaluate object discovery performance with our proposals on four datasets: Object Discovery (OD), VOC\_6x2, VOC\_all and VOC12. OD is a small dataset with three classes \textit{airplane}, \textit{car} and \textit{horse}, and 100 images per class, among which 18, 11 and 7 images are outliers (images not including an object of the corresponding class) respectively. VOC\_all is a subset of the PASCAL VOC 2007 dataset~\cite{pascal-voc-2007}  obtained by eliminating all images containing only \textit{difficult} or \textit{truncated} objects as well as \textit{difficult} or \textit{truncated} objects in retained images. It has 3550 images and 6661 objects. VOC\_6x2 is a subset of VOC\_all which contains images of 6 classes \textit{aeroplane}, \textit{bicycle}, \textit{boat}, \textit{bus}, \textit{horse} and \textit{motorbike} divided into 2 views \textit{left} and \textit{right}. In total, VOC\_6x2 contains 463 images of 12 classes. VOC12 is a subset of the PASCAL VOC 2012 dataset~\cite{pascal-voc-2012} and obtained in the same way as VOC\_all. It contains 7838 images and figures 13957 objects. For large-scale experiments, we randomly choose 20000 images from the training set of COCO~\cite{Lin2014cocodataset} and eliminate those containing only \textit{crowd} bounding boxes as well as bounding boxes marked as \textit{crowd} in retained images. The resulting dataset, which we call COCO\_20k, has 19817 images and 143951 objects.

As single-object discovery and colocalization performance measure, we use \textit{correct localization} (CorLoc) defined as the percentage of images correctly localized. In our context, this means the intersection over union ($IoU$) between one of the ground-truth regions and one of the predicted regions in the image is greater than 0.5. Since CorLoc does not take into account multiple detections per image, for multi-object discovery, we use instead \textit{detection rate} at the IoU threshold of $0.5$ as measure of performance. Given some threshold $\zeta$, detection rate at $IoU=\zeta$ is the percentage of ground-truth bounding boxes that have an \textit{IoU} with one of the retained proposals greater than $\zeta$. We run the experiments in both the colocalization setting, where the algorithm is run separately on each class of the dataset, and the average CorLoc/detection rate over all classes is computed as the overall performance measure on the dataset, and the true \textit{discovery} setting where the whole dataset is considered as a single class.

\subsection{Implementation Details}
\label{sec:implementation}

\noindent{\textbf{Features.}} We test our methods with the pre-trained CNN features from VGG16 and VGG19~\cite{Symonian2014verydeep}. For generating region proposals, we apply the algorithm described in Section~\ref{sec:unsup_region_proposals} separately to the layers right before the last two max pooling layers of the networks (\textit{relu4\_3} and \textit{relu5\_3} in VGG16, \textit{relu4\_4} and \textit{relu5\_4} in VGG19), then fuse proposals generated from the two layers as our final set of proposals. Note that using CNN features at multiple layers is important as different layers capture different visual patterns in images~\cite{zeiler2013understanding}. One could also use more layers from VGG16 (e.g., layers \textit{relu3\_3}, \textit{relu4\_2} or \textit{relu5\_2}) but we only use two for the sake of efficiency. In experiments with OSD, we extract features for the region proposals by applying the RoI pooling operator introduced in Fast-RCNN~\cite{girshickICCV15fastrcnn} to layer \textit{relu5\_3} of VGG16.

\noindent{\textbf{Region Proposal Generation Process.}} For finding robust local maxima of the global saliency maps $s_g$, we rank its locations using persistence~\cite{Chazal2013persistence,Edelsbrunner2009introtopo,Edelsbrunner2002topo,Oudot2015persistence,Zomorodian2005compute}. Concretely, we consider $s_g$ as a 2D image and each location in it as a pixel. We associate with each pixel a cluster (the 4-neighborhood connected component of pixels that contains it), together with a ``birth'' (its own saliency) and ``death'' time (the highest value for which one of the pixels in its cluster also belongs to the cluster of a pixel with higher saliency, or, if no such location exists, the lowest saliency value in the map). The persistence of a pixel is defined as the difference between its birth and death times. A sorted list of pixels in decreasing persistence order is computed, and the local maxima are chosen as the top pixels in the list. For additional robustness, we also apply non maximum suppression on the list over a $3\times 3$ neighborhood. Since the saliency map created from CNN feature maps can be very noisy, we eliminate locations with score in $s_g$ below $\alpha \max{s_g}$ before computing the persistence to obtain only good local maxima. We also eliminate locations with score smaller than the average score in $s_y$ and whose score in $s_g$ is smaller than $\beta$ times the average score in $s_g$. We choose the value of the pair $(\alpha, \beta)$ in $\{0.3, 0.5\} \times \{0.5, 1\}$ by conducting small-scale object discovery on VOC\_6x2. We find that $(\alpha, \beta) = (0.3, 0.5)$ yields the best performance and gives local saliency maps that are not fragmented while eliminating well irrelevant locations across settings and datasets. We take up to $20$ local maxima (after non-maximum suppression) and use $50$ linearly spaced thresholds between the lowest and the highest scores in each local saliency map to generate proposals. 

\noindent{\textbf{Object Discovery Experiments.}} 
For single-object colocalization and discovery, following~\cite{Vo2019UnsupOptim}, we use $\nu=5, \tau=10$ and apply the OSD's post processing to obtain the final localization result. For multi-object setting, we use $\nu=50$, $\tau=10$ and apply the post processing with non-maximum suppression at $IoU=0.7$ to retain at most 5 regions in the final result. 
On large classes/datasets, we pre-filter the set of neighbors that are considered in the optimization for each image, using the cosine similarity between features from the fully connected layer \textit{fc6} of the pre-trained network, following~\cite{Babenko2014neuralcode}. The number of potential neighbors of each image is fixed to $50$ in all experiments where the pre-filtering is necessary. 

\begin{figure*}[tb]
	\begin{center}
	    \begin{subfigure}{0.24\textwidth}
			\includegraphics[width=\linewidth]{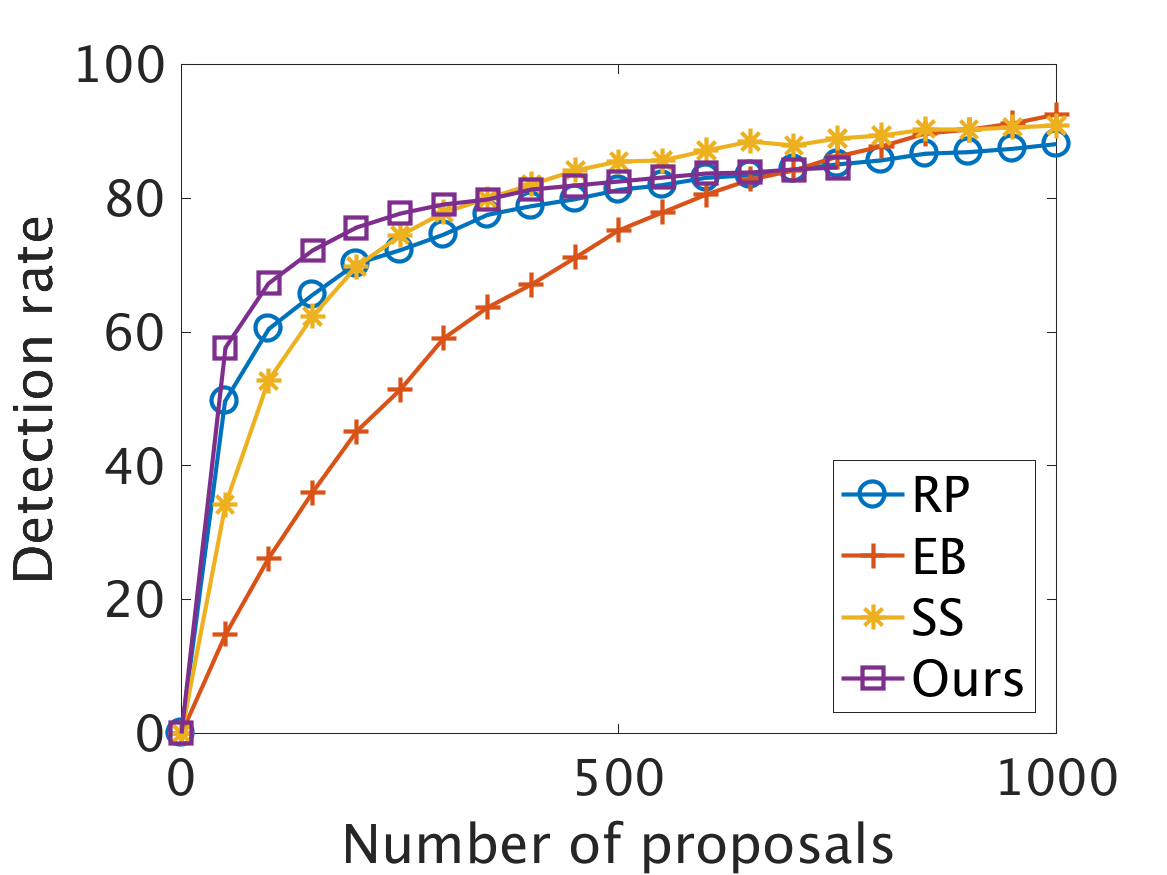}
			\caption{\small $IoU=0.5$.}
			\label{fig:iou05}
		\end{subfigure}
		\begin{subfigure}{0.24\textwidth}
			\includegraphics[width=\linewidth]{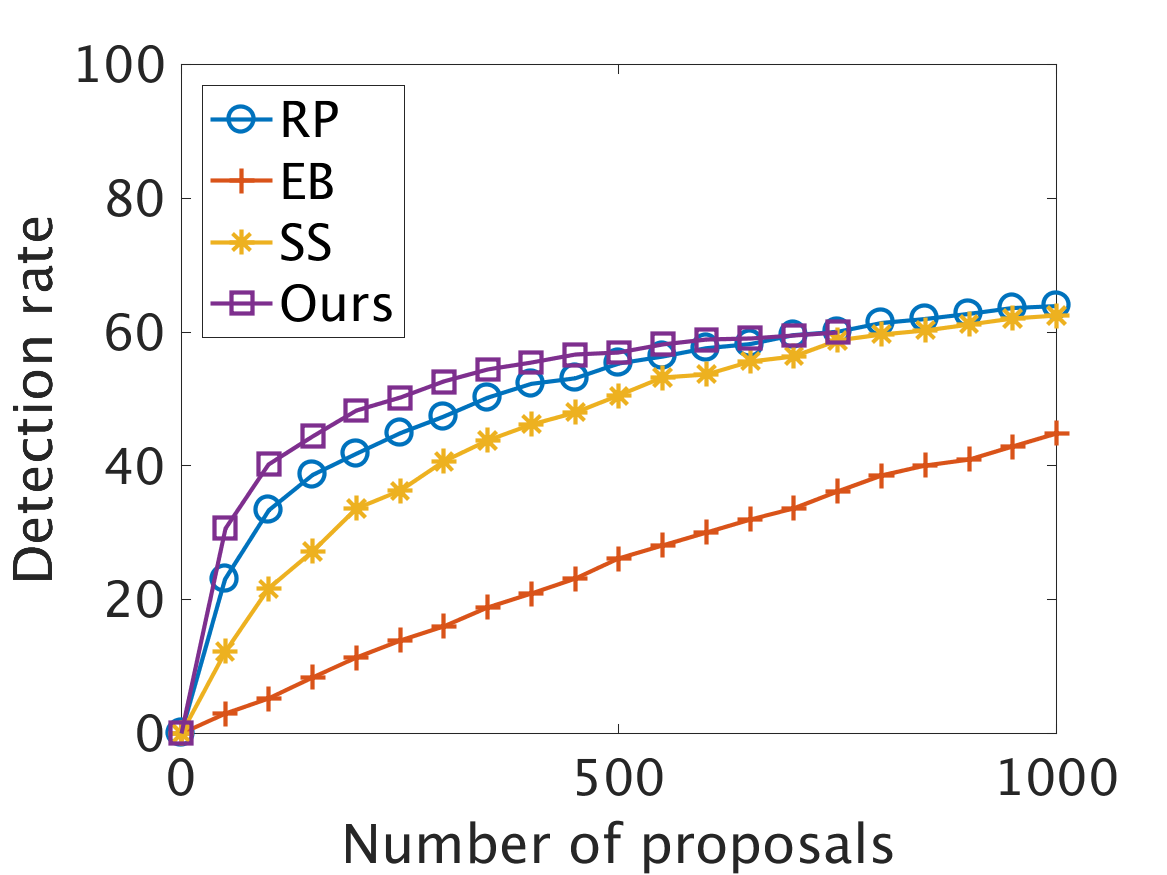}
			\caption{\small $IoU=0.7$.}
			\label{fig:iou07}
		\end{subfigure}
		\begin{subfigure}{0.24\textwidth} \includegraphics[width=\linewidth]{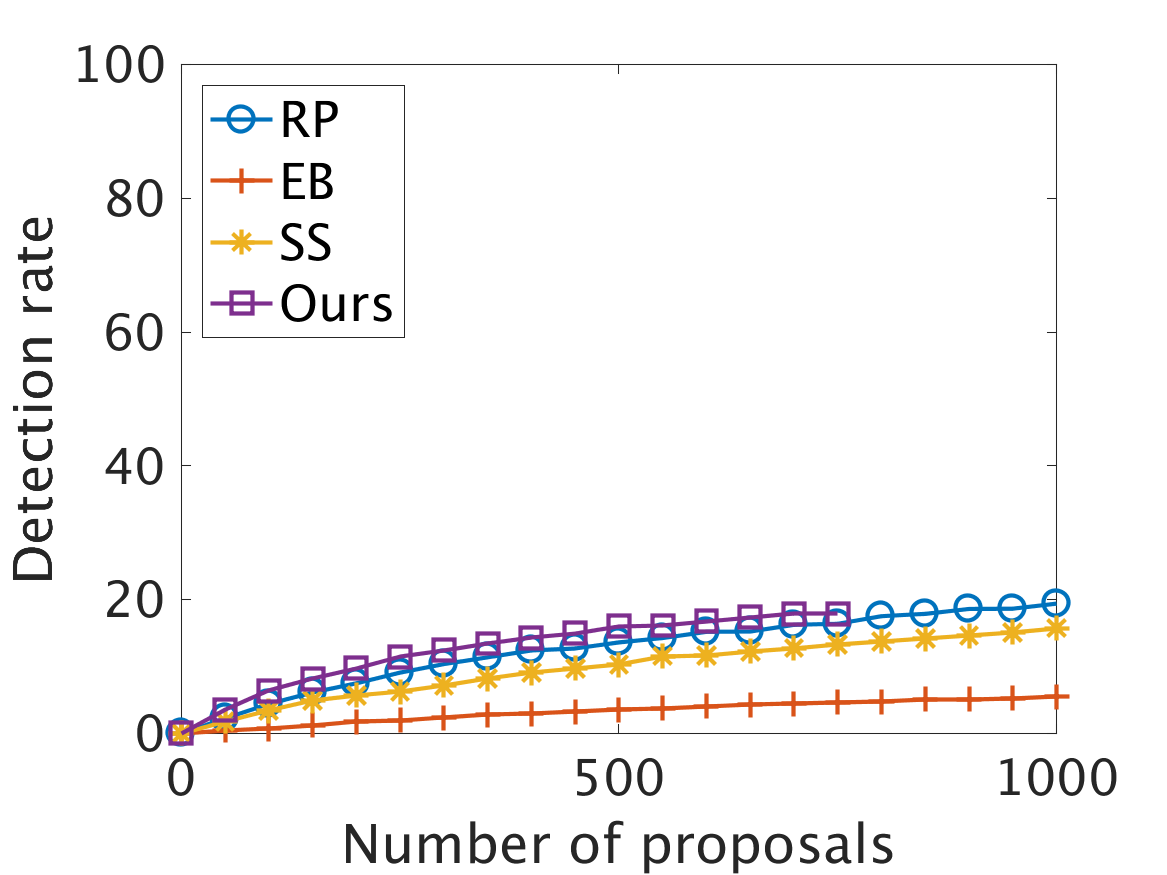}
			\caption{\small $IoU=0.9$.}
			\label{fig:iou09}
		\end{subfigure}
		\begin{subfigure}{0.24\textwidth}
			\includegraphics[width=\linewidth]{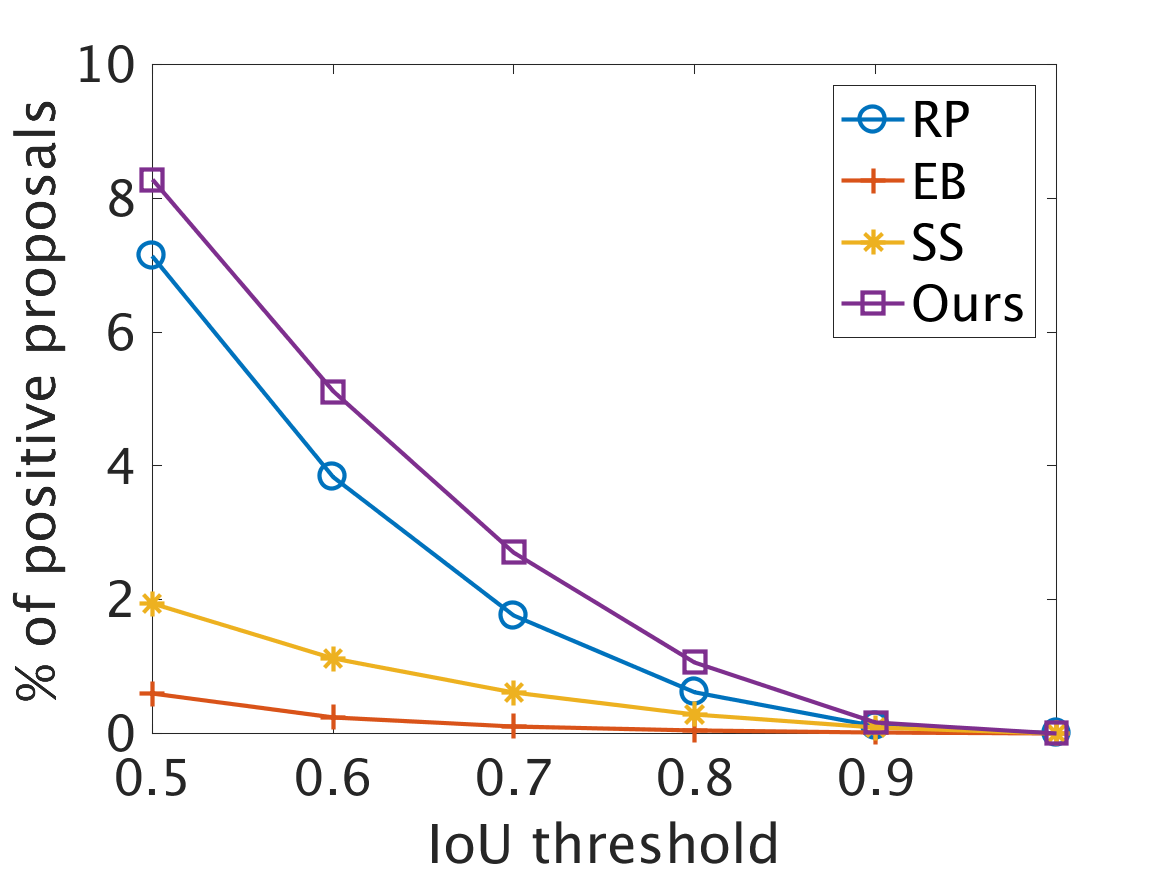}
			\caption{\small positive regions.}
			\label{fig:percentage_positive_proposals}
		\end{subfigure}
	\end{center}
	\caption{\small Quality of proposals by different methods. (\protect\subref{fig:iou05}-\protect\subref{fig:iou09}): Detection rate by number of proposals at different $IoU$ thresholds of randomized Prim (RP)~\cite{Manen2013prim}, edgeboxes (EB)~\cite{zitnick2014edge}, selective search (SS)~\cite{uijlings2013selective} and ours; (\protect\subref{fig:percentage_positive_proposals}): Percentage of positive proposals for the four methods.}
	\label{fig:quality_of_proposals}
\end{figure*}

\subsection{Region Proposal Evaluation}
Following other works on region proposals~\cite{Manen2013prim,uijlings2013selective,zitnick2014edge}, we evaluate the quality of our proposals on PASCAL VOC 2007 using the \textit{detection rate} at various \textit{IoU} thresholds. But since we intend to later use our proposals for object discovery, unlike other works, we evaluate directly our proposals on VOC\_all instead of the test set of VOC 2007 to reveal the link between the quality of proposals and the object discovery performance. Figure~\ref{fig:quality_of_proposals}(\protect\subref{fig:iou05}-\protect\subref{fig:iou09}) shows the performance of different proposals on VOC\_all. It can be seen that our method performs better than others at a very high overlap threshold ($0.9$) regardless of the number of proposals allowed. At medium threshold ($0.7$), our proposals are on par (or better for fewer than 500 proposals) with those from selective search~\cite{uijlings2013selective} and randomized Prim~\cite{Manen2013prim} and much better than those from edgeboxes~\cite{zitnick2014edge}. At a small threshold ($0.5$), our method is still on par with randomized Prim and edgeboxes, but does not fare as well as selective search. It should be noted that randomized Prim is supervised whereas the others are unsupervised.

In OSD, localizing an object in an image means singling out a 
{\em positive} proposal, that is, a proposal having an \textit{IoU} greater than some threshold with object bounding boxes. It is therefore easier to localize the object if the percentage of positive region proposals is larger. As shown by Fig.~\ref{fig:quality_of_proposals}(\protect\subref{fig:percentage_positive_proposals}), our method performs very well according to this criterion: Over 8\% of our proposals are positive at an \textit{IoU} threshold of $0.5$, and over 3\% are still positive for an \textit{IoU} of $0.7$. Also, randomized Prim and our method are by far better than selective search and edgeboxes, which explains the superior object discovery performance of the former over the latter (\textit{cf}.~\cite{Vo2019UnsupOptim} and Table~\ref{table:separate_compare_proposals}). Note that region proposals with a high percentage of positive ones could also be used in other tasks, i.e., weakly supervised object detection, but this is left for future work.

\subsection{Object Discovery Performance}
\noindent{\textbf{Single-Object Colocalization and Discovery.}} An important component of OSD is the similarity model used to compute score matrices $S_{ij}$, which, in~\cite{Vo2019UnsupOptim}, is the Probabilistic Hough Matching (PHM) algorithm~\cite{CKSP15}. Vo {\em et al.}~\cite{Vo2019UnsupOptim} introduce two scores, \textit{confidence} score and \textit{standout} score, but use only the latter for it gives better performance. Since our new proposals come with different statistics, we test both scores in our experiments. Table~\ref{table:separate_confidence_standout} compares colocalization performance on OD, VOC\_6x2 and VOC\_all of OSD using the confidence and standout scores as well as our proposals. It can be seen that on VOC\_6x2 and VOC\_all, the confidence score does better than the standout score, while on OD, the latter does better. This is in fact not particularly surprising since images in OD generally contain bigger objects (relative to image size) than those in the other datasets. In fact, although the standout score is used on all datasets in~\cite{CKSP15} and~\cite{Vo2019UnsupOptim}, the authors adjust the parameter $\gamma$ (see~\cite{CKSP15}) used in computing their standout score to favor larger regions when running their models on OD. In all of our experiments from now on, we use the standout score on OD and the confidence score on other datasets (VOC\_6x2, VOC\_all, VOC12 and COCO\_20k).
\begin{table}[tb]
    \parbox[t][][t]{0.38\linewidth}{
    \caption{\small Colocalization performance with our proposals in different configurations of OSD}
    \label{table:separate_confidence_standout}
    \centering
    \resizebox{0.38\textwidth}{!}{%
    \setlength{\tabcolsep}{0.5em}
    \begin{tabular}{ccc}
        \toprule
        Config. & Confidence & Standout  \\
        \midrule
        OD & 83.7 $\pm$ 0.4 & \textbf{89.0 $\pm$ 0.6} \\
        VOC\_6x2 & \textbf{73.6 $\pm$ 0.6} & 64.1 $\pm$ 0.3 \\
        VOC\_all & \textbf{44.7 $\pm$ 0.3} & 41.4 $\pm$ 0.1 \\
        \bottomrule
    \end{tabular}
    }
    }
    \hfill 
    \parbox[t][][t]{0.6\linewidth}{
    \caption{\small Colocalization performance for different values of hyper-parameters}
    \label{table:hyperparameters}
    \centering
    \resizebox{0.6\textwidth}{!}{%
    \setlength{\tabcolsep}{0.5em}
    \begin{tabular}{ccccc}
        \toprule
        ($u$, $v$) & (20,50) & (20,100) & (50,50) & (50,100)  \\
        \midrule
        CorLoc & 73.6 $\pm$ 0.8 & 73.4 $\pm$ 0.7 & 73.3 $\pm$ 1.1 & 74.2 $\pm$ 0.8 \\
        $p$ & 760 & 882 & 1294 & 1507 \\
        \bottomrule
    \end{tabular}
    }
    }
\end{table}

Our proposal generation process introduces a few hyper-parameters. Apart from $\alpha$ and $\beta$, two other important hyper-parameters are the number of local maxima $u$ and the number of thresholds $v$ which together control the number of proposals $p$ per image returned by the process. We study their influence on the colocalization performance by conducting experiments on VOC\_6x2 and report the results in Table~\ref{table:hyperparameters}. It shows that the colocalization performance does not depend much on the values of these parameters. Using ($u=50, v=100$) actually gives the best performance but with twice as many proposals as ($u=20, v=50$). For efficiency, we use $u=20$ and $v=50$ in all of our experiments.

We report in Table~\ref{table:separate_compare_proposals} the performance of OSD and rOSD on OD, VOC\_6x2 and VOC\_all with different types of proposals. It can be seen that our proposals give the best results on all datasets among all types of proposals with significant margins: 6.1\%, 2.1\% and 3.0\% in colocalization and 5.3\%, 0.5\% and 4.7\% in discovery, respectively. It is also noticeable that our proposals not only fare much better than the unsupervised ones (selective search and edgeboxes) but outperform those generated by randomized Prim, an algorithm trained with bounding box annotation.

\begin{table}[tb]
    \caption{\small Single-object colocalization and discovery performance of OSD with different types of proposals. We use VGG16 features to represent regions in these experiments}
    \label{table:separate_compare_proposals}
    \centering
    \resizebox{0.9\textwidth}{!}{%
    \setlength{\tabcolsep}{0.5em}
    \begin{tabular}{ccccccc}
        \toprule
        \multirow{2}{*}{Region proposals} & \multicolumn{3}{c}{Colocalization} & \multicolumn{3}{c}{Discovery} \\ 
          & OD & VOC\_6x2 & VOC\_all & OD & VOC\_6x2 & VOC\_all \\
        \midrule
        Edgeboxes~\cite{zitnick2014edge} & 81.6 $\pm$ 0.3 & 54.2 $\pm$ 0.3 & 29.7 $\pm$ 0.1 & 81.4 $\pm$ 0.3 & 55.2 $\pm$ 0.3 & 32.6 $\pm$ 0.1 \\
        Selective search~\cite{uijlings2013selective} & 82.2 $\pm$ 0.2 & 54.5 $\pm$ 0.3 & 30.9 $\pm$ 0.1 & 81.3 $\pm$ 0.3 & 57.8 $\pm$ 0.2 & 33.0 $\pm$ 0.1 \\
        Randomized Prim~\cite{Manen2013prim} & 82.9 $\pm$ 0.3 & 71.5 $\pm$ 0.3 & 42.8 $\pm$ 0.1 & 82.5 $\pm$ 0.1 & \underline{70.6 $\pm$ 0.4} & 44.5 $\pm$ 0.1 \\
        \midrule
        Ours (OSD) & \textbf{89.0 $\pm$ 0.6} & \textbf{73.6 $\pm$ 0.6} & \underline{44.7 $\pm$ 0.3} & \textbf{87.8 $\pm$ 0.4} & 69.2 $\pm$ 0.5 & \underline{48.7 $\pm$ 0.3} \\
        Ours (rOSD) & \textbf{89.0 $\pm$ 0.5} & \underline{73.3 $\pm$ 0.5} & \textbf{45.8 $\pm$ 0.3} & \underline{87.6 $\pm$ 0.3} & \textbf{71.1 $\pm$ 0.8} & \textbf{49.2 $\pm$ 0.2} \\       
        \bottomrule
    \end{tabular}
    }
\end{table}

We compare OSD and rOSD using our region proposals to the state of the art in Table~\ref{table:single_separate} (colocalization) and Table~\ref{table:single_mixed} (discovery). In their experiments, Wei {\em et al.}~\cite{Wei2019ddtplus} only use features from VGG19. We have conducted experiments with features from both VGG16 and VGG19 but only present experiment results with VGG19 features in comparisons with~\cite{Wei2019ddtplus} due to the space limit. A more comprehensive comparison with features from VGG16 is included in the supplementary material. It can be seen that our use of CNN features (for both creating proposals and representing them in OSD) consistently improves the performance compared to the original OSD~\cite{Vo2019UnsupOptim}. It is also noticeable that rOSD performs significantly better than OSD on the two large datasets (VOC\_all and VOC12) while on the two smaller ones (OD and VOC\_6x2), their performances are comparable. It is due to the fact that images in OD and VOC\_6x2 mostly contain only one well-positioned object thus bad local maxima are not a big problem in the optimization while images in VOC\_all and VOC12 contain much more complex scenes and the optimization works better with more regularization. In overall, we obtain the best results on the two smaller datasets, fare better than~\cite{Li2016mimick} but are behind~\cite{Wei2019ddtplus} on VOC\_all and VOC12 in the colocalization setting. It should be noticed that while methods for image colocalization~\cite{Li2016mimick,Wei2019ddtplus} suppose that images in the collection come from the same category and explicitly exploit this assumption, rOSD is intended to deal with the much more difficult and general object discovery task. Indeed, in discovery setting, rOSD outperforms~\cite{Wei2019ddtplus} by a large margin, 5.9\% and 4.9\% respectively on VOC\_all and VOC12.

\begin{table*}[tb]
    \centering
    \caption{\small Single-object colocalization performance of our approach compared to the state of the art. Note that Wei {\em et al.} ~\cite{Wei2019ddtplus} outperform our method on VOC\_all and VOC12 in this case, but the situation is clearly reversed in the much more difficult discovery setting, as demonstrated in Table~\ref{table:single_mixed}}
    \label{table:single_separate}
    \resizebox{0.8\textwidth}{!}{%
    \setlength{\tabcolsep}{1em}
    \begin{tabular}{cccccc}
        \toprule
        Method & Features & OD & VOc\_6x2 & VOC\_all & VOC12 \\
        \midrule
        Cho {\em et al.}~\cite{CKSP15} & WHO & 84.2 & 67.6 & 37.6 & - \\
        Vo {\em et al.}~\cite{Vo2019UnsupOptim} & WHO & 87.1 $\pm$ 0.5 & 71.2 $\pm$ 0.6 & 39.5 $\pm$ 0.1 & - \\
        \midrule
        Li {\em et al.}~\cite{Li2016mimick} & VGG19 & - & - & 41.9 & 45.6 \\
        Wei {\em et al.}~\cite{Wei2019ddtplus} & VGG19 & 87.9 & 67.7 & \textbf{48.7} & \textbf{51.1} \\
        Ours (OSD) & VGG19 & \textbf{90.3 $\pm$ 0.3} & 75.3 $\pm$ 0.7 & 45.6 $\pm$ 0.3 & 47.8 $\pm$ 0.2 \\
        Ours (rOSD) & VGG19 & 90.2 $\pm$ 0.3 & \textbf{76.1 $\pm$ 0.7} & 46.7 $\pm$ 0.2 & 49.2 $\pm$ 0.1 \\
        \bottomrule
    \end{tabular}
    }
\end{table*}

\begin{table*}[tb]
    \centering
    \caption{\small Single-object discovery performance on the datasets with our proposals compared to the state of the art}
    \label{table:single_mixed}
    \resizebox{0.8\textwidth}{!}{%
    \setlength{\tabcolsep}{1em}
    \begin{tabular}{cccccc}
        \toprule
        Method  & Features & OD & VOC\_6x2 & VOC\_all & VOC12 \\ 
        \midrule
        Cho {\em et al.}~\cite{CKSP15} & WHO & 82.2 & 55.9 & 37.6 & -\\ 
        Vo {\em et al.}~\cite{Vo2019UnsupOptim} & WHO & 82.3 $\pm$ 0.3 & 62.5 $\pm$ 0.6 & 40.7 $\pm$ 0.2 & - \\
        \midrule
        Wei {\em et al.}~\cite{Wei2019ddtplus} & VGG19 & 75.0 & 54.0 & 43.4 & 46.3 \\
        Ours (OSD) & VGG19 & 89.1 $\pm$ 0.4 & 71.9 $\pm$ 0.7 & 47.9 $\pm$ 0.3 & 49.2 $\pm$ 0.2 \\
        Ours (rOSD) & VGG19 & \textbf{89.2 $\pm$ 0.4} & \textbf{72.5 $\pm$ 0.5} & \textbf{49.3 $\pm$ 0.2} & \textbf{51.2 $\pm$ 0.2} \\
        \bottomrule
    \end{tabular}
    }
\end{table*}

\begin{figure*}[h]
\centering
\parbox[t][][t]{0.85\linewidth}{
\includegraphics[width=\linewidth]{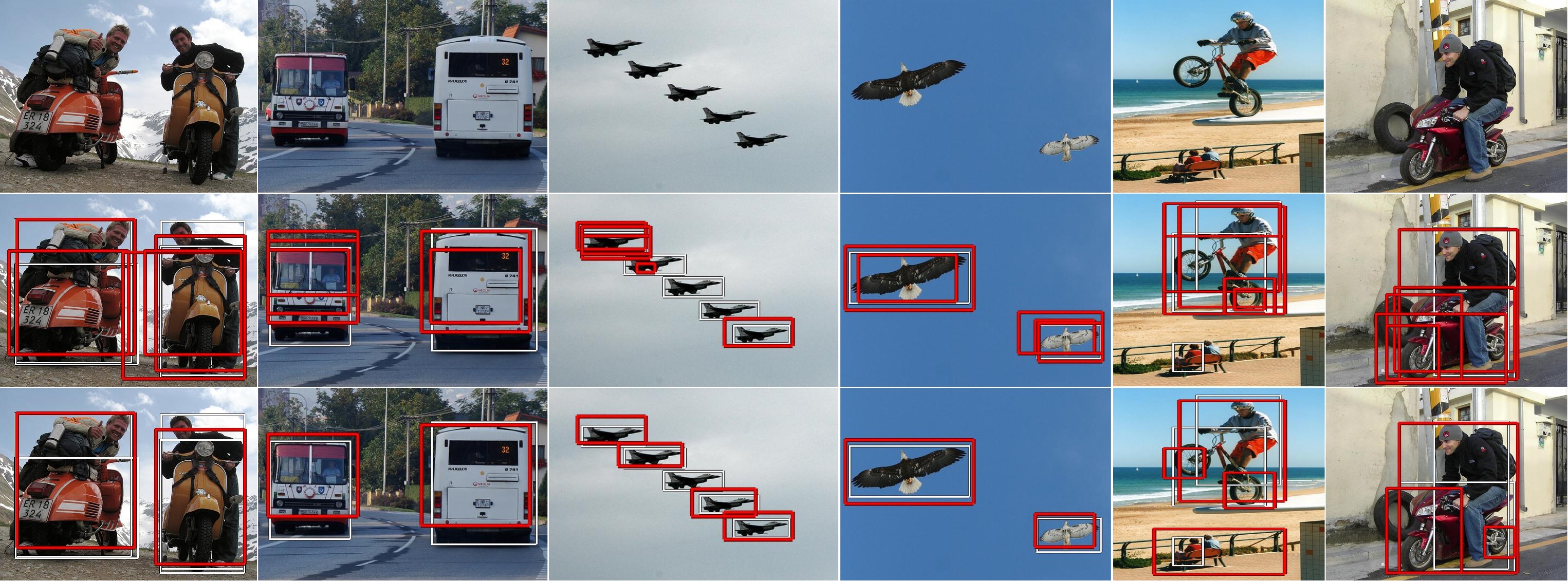}
}
\vspace{-5pt}
\caption{\small Qualitative multi-object discovery results obtained with rOSD. White boxes are ground truth objects and red ones are our predictions. Original images are in the first row. Results with $\nu=50$ and $IoU=0.7$ are in the second row. Results with $\nu=25$ and $IoU=0.3$ are in the third row.}
\label{fig:teaser}
\end{figure*}

\noindent{\textbf{Multi-Object Colocalization and Discovery.}}
We demonstrate the effectiveness of rOSD in multi-object colocalization and discovery on VOC\_all and VOC12 datasets, which contain images with multiple objects.  We compare the performance of OSD and rOSD to Wei {\em et al.}~\cite{Wei2019ddtplus} in Table~\ref{table:multiple_perf_all}. Although~\cite{Wei2019ddtplus} tackles only the single-object colocalization problem, we modify their method to have a reasonable baseline for the multi-object colocalization and discovery problem. Concretely, we take the bounding boxes around the 5 largest connected components of positive locations in the image's \textit{indicator matrix}~\cite{Wei2019ddtplus} as the localization results. It can be seen that our method obtains the best performance with significant margins to the closest competitor across all datasets and settings. It is also noticeable that rOSD, again, significantly outperforms OSD in this task. 
An illustration of multi-object discovery is shown in Fig.~\ref{fig:teaser}. For a fair comparison, we use high values of $\nu$ ($50$) and $IoU$ ($0.7$) in the multi-object experiments to make sure that both OSD and rOSD return approximately 5 regions per image. Images may of course contain fewer than 5 objects. In such cases, OSD and rOSD usually return overlapping boxes around the actual objects. We can often eliminate these overlapping boxes and obtain better qualitative results by using smaller $\nu$ and $IoU$ threshold values. It can be seen in Fig.~\ref{fig:teaser} that with $\nu=25$ and $IoU=0.3$, rOSD is able to return bounding boxes around objects without many overlapping regions. Note however that the quantitative results may worsen due to the reduced number of regions returned and the fact that many images contain objects that highly overlap, e.g., the last two columns of Fig.~\ref{fig:teaser}. In such cases, a small $IoU$ threshold prevents discovering all of these objects. See supplementary document for more visualizations and details.

\begin{table}[tb]
    \centering
    \caption{\small Multi-object colocalization and discovery performance of rOSD compared to competitors on VOC\_all and VOC12 datasets}
    \label{table:multiple_perf_all}
    \resizebox{0.8\textwidth}{!}{%
    \setlength{\tabcolsep}{1em}
    \begin{tabular}{cccccc}
        \toprule
        \multirow{2}{*}{Method} & \multirow{2}{*}{Features} & \multicolumn{2}{c}{Colocalization} & \multicolumn{2}{c}{Discovery} \\
         & & VOC\_all & VOC12 & VOC\_all & VOC12 \\
        \midrule
        Vo {\em et al.}~\cite{Vo2019UnsupOptim} & WHO & 40.7 $\pm$ 0.1 & - & 30.7 $\pm$ 0.1 & - \\
        \midrule
        Wei {\em et al.}~\cite{Wei2019ddtplus} & VGG19 & 43.3 & 45.5 & 28.1 & 30.3 \\
        Ours (OSD) & VGG19 & 46.8 $\pm$ 0.1 & 47.9 $\pm$ 0.0 & 34.8 $\pm$ 0.0 & 36.8 $\pm$ 0.0 \\
        Ours (rOSD) & VGG19 & \textbf{49.4 $\pm$ 0.1} & \textbf{51.5 $\pm$ 0.1} & \textbf{37.6 $\pm$ 0.1} & \textbf{40.4 $\pm$ 0.1} \\
        \bottomrule
    \end{tabular}
    }
\end{table}

\noindent{\textbf{Large-Scale Object Discovery.}}
\label{sec:result_large_scale}
We apply our large-scale algorithm in the discovery setting on VOC\_all, VOC12 and COCO\_20k which are randomly partitioned respectively into $5$, $10$ and $20$ parts of roughly equal sizes. In the first stage of all experiments, we prefilter the initial neighborhood of images and keep only $50$ potential neighbors. We choose $\nu=50$ and keep $K_1$ (which are $250$, $500$ and $1000$ respectively on VOC\_all, VOC12 and COCO\_20k) positive entries in each score matrix. In the second stage, we run rOSD (OSD) on the entire datasets with $\nu=5$, limit the number of potential neighbors to $50$ and use score matrices with only $50$ positive entries. We choose $K_1$ such that each run in the first stage and the OSD run in the second stage have the same memory cost, hence the values of $K$ chosen above. As baselines, we have applied rOSD (OSD) directly to the datasets, keeping $50$ positive entries (baseline 1) and $1000$ positive entries (baseline 2) in score matrices. Table~\ref{table:large_scale_performance} shows the object discovery performance on VOC\_all, VOC12 and COCO\_20k for our large-scale algorithm compared to the baselines. It can be seen that our large-scale two-stage rOSD algorithm yields significant performance gains over the baseline 1, obtains an improvement of 6.6\%, 9.3\% and 4.0\% in single-object discovery and 2.9\%, 4.0\% and 0.4\% in multi-object discovery, respectively on VOC\_all, VOC12 and COCO\_20k. Interestingly, large-scale rOSD also outperforms the baseline 2, which has a much higher memory cost, on VOC\_all and VOC12.

\begin{table}[tb]
    \centering
    \caption{\small Performance of our large-scale algorithm compared to the baselines. Our method and baseline 1 have the same memory cost, which is much smaller than the cost of baseline 2 . Also, due to memory limits, we cannot run baseline 2 on COCO\_20k}
    \label{table:large_scale_performance}
    \resizebox{0.9\textwidth}{!}{%
    \setlength{\tabcolsep}{0.5em}
    \begin{tabular}{ccccccc}
        \toprule
        \multirow{2}{*}{Method} & \multicolumn{3}{c}{Single-object} & \multicolumn{3}{c}{Multi-object} \\
          & VOC\_all & VOC12 & COCO\_20k & VOC\_all & VOC12 & COCO\_20k \\
        \midrule
        Baseline 1 (OSD) & 41.1 $\pm$ 0.3 & 40.5 $\pm$ 0.2 & 43.6 $\pm$ 0.2 & 31.4 $\pm$ 0.1 & 32.4 $\pm$ 0.0 & 10.5 $\pm$ 0.0 \\
        Baseline 1 (rOSD) & 42.8 $\pm$ 0.3 & 42.6 $\pm$ 0.2 & 44.5 $\pm$ 0.1 & 35.4 $\pm$ 0.2 & 37.2 $\pm$ 0.1 & 11.6 $\pm$ 0.0 \\
        \midrule
        Baseline 2 (OSD) & 47.9 $\pm$ 0.3 & 49.2 $\pm$ 0.2 & - & 34.8 $\pm$ 0.0 & 36.8 $\pm$ 0.0 & - \\
        Baseline 2 (rOSD) & 49.3 $\pm$ 0.2 & 51.2 $\pm$ 0.2 & - & 37.6 $\pm$ 0.1 & 40.4 $\pm$ 0.1 & - \\
        \midrule
        Large-scale OSD & 45.5 $\pm$ 0.3 & 46.3 $\pm$ 0.2 & 46.9 $\pm$ 0.1 & 34.6 $\pm$ 0.0 & 36.9 $\pm$ 0.0 & 11.1 $\pm$ 0.0 \\
        Large-scale rOSD & \textbf{49.4 $\pm$ 0.1} & \textbf{51.9 $\pm$ 0.1} & \textbf{48.5 $\pm$ 0.1} & \textbf{38.3 $\pm$ 0.0} & \textbf{41.2 $\pm$ 0.1} & \textbf{12.0 $\pm$ 0.0} \\
        \bottomrule
    \end{tabular}
    }
\end{table}

\noindent{\textbf{Execution time.}}
Similar to~\cite{Vo2019UnsupOptim}, our method requires computing the similarity scores for a large number of image pairs which makes it computationally costly. It takes in total $478$ paralellizable CPU hours, $300$ unparallelizable CPU seconds and $1$ GPU hour to run single-object discovery on VOC\_all with $3550$ images. This is more costly compared to only $812$ GPU seconds needed by DDT+~\cite{Wei2019ddtplus} but is less costly than~\cite{Vo2019UnsupOptim} using CNN features. The latter requires $546$ paralellizable CPU hours, $250$ unparalellizable CPU seconds and 4 GPU hours. Note that the unparallelizable computational cost, which comes from the main OSD algorithm, grows very fast (at least linearly in theory, it takes 2.3 hours on COCO\_20k in practice) with the data set's size and is the time bottleneck in large scale.

\section{Conclusion}
We have presented an unsupervised algorithm for generating region proposals from CNN features trained on an auxiliary and unrelated task. Our proposals come with an intrinsic structure which can be leveraged as an additional regularization in the OSD framework of Vo {\em et al.}~\cite{Vo2019UnsupOptim}. The combination of our proposals and regularized OSD gives comparable results to the current state of the art in image colocalization, sets a new state-of-the-art single-object discovery and has proven effective in the multi-object discovery. We have also successfully extended OSD to the large-scale case and show that our method yields significantly better performance than plain OSD. Future work will be dedicated to investigating other applications of our region proposals. 

\noindent{\textbf{Acknowledgments.}} This work was supported in part by the Inria/NYU collaboration, the Louis Vuitton/ENS chair on artificial intelligence and the French government under management of Agence Nationale de la Recherche as part of the “Investissements d’avenir” program, reference ANR19-P3IA-0001 (PRAIRIE 3IA Institute). Huy V. Vo was supported in part by a Valeo/Prairie CIFRE PhD Fellowship.

\clearpage
%
%
 \ifx\URL\undefined \def\URLset#1{{\tt #1}\catcode`\~=\active\catcode`\_=8}
  \def\URL{\catcode`\~=12 \catcode`\_=12 \URLset} \fi

\clearpage
\begin{center}
    \textbf{\large Supplementary materials: Toward unsupervised, multi-object discovery in large-scale image collections}
\end{center}
\setcounter{section}{0}
\setcounter{table}{0}
\setcounter{figure}{0}
\section{Regularized OSD (rOSD)}
We have presented in the paper a new version of the OSD formulation~\cite{Vo2019UnsupOptim} with added constraints based on the structure of our region proposals. Concretely, we propose to solve the optimization problem:
\begin{equation}
\!\!\!\!\!\!\underset{x,e}{\mathrm{max}}  
S(x,e)=\!\!\sum_{i=1}^n \!\sum\limits_{j \in N(i)}\!\!\! e_{ij} x_i^T S_{ij} x_j, \ 
\text{s.t.}\forall i\ \left\{\begin{array}{l}
\ \sum\limits_{k=1}^p x_i^k\le \nu,\\
\!\sum\limits_{k \in G_{ig}} x_i^k \le 1, \text{ for all groups } g\ \\
\ \sum\limits_{j\neq i} e_{ij}\le\tau.
\end{array}\right.
\label{eq:rOSD}
\end{equation}

We solve this problem with an iterative block-coordinate ascent algorithm similar to OSD. Its iterations are illustrated in Algorithm~\ref{algo:rOSD}.

\begin{algorithm}[htb]
\SetAlgoLined
\KwResult{A solution to rOSD.}
Input: $G_i$, $\nu$, $\tau$, $S_{ij}$, number $n$ of images.\\
Initialization: $x_i = \mathbbold{1}_p\ \forall i$, $e_{ij} = 1\ \forall i \neq j$.\\
\For{$i = 1$ \text{to} $n$}{
  Compute the vector $R$ containing the scores of regions in image $i$. \\ 
  $R \longleftarrow \sum_{j \neq i}^n (e_{ij}S_{ij} + e_{ji}S_{ji}^T)x_j.$ \\
  $I \longleftarrow \emptyset$. \\
  \For{$g=1;\ g \leq L_i$}{
    Find the region $g^*$ with highest score $R(g^*)$ in the group $G_{ig}$. \\
    $I \longleftarrow I \cup \{g^*\}$.\\
  }
  Choose $\nu$ regions in $I$ with highest scores in $R$, assign their corresponding variables to $1$. Assign the variables of other regions to $0$.
}
\For{$i=1$ to $n$}{
  Compute the indices $j_1$ to $j_\tau$ of the $\tau$ largest scalars $x_{i}^T S_{ij} x_{j}\ (1\leq j\leq n)$.\\
  $e_i\longleftarrow 0$.\\
  \For{$t=1;\ t \leq \tau$}{
    $e_{ij_t}\longleftarrow 1$.\\
  }
}
\caption{Block coordinate ascent algorithm for rOSD.}
\label{algo:rOSD}
\end{algorithm}
Note that the output of Algorithm~\ref{algo:rOSD} depends on the order in which the variables $x_i$ are processed in its first \textit{for} loop. In our implementation, we use a \textit{different} random permutation of $(1,...,n)$ in each iteration of the optimization. For each experiment, we run rOSD several times and report the average performance of all runs as the final performance.

\section{Large-Scale Object Discovery Algorithm}
We summarize in Algorithm~\ref{algo:large_scale_OSD} our proposed large-scale algorithm for object discovery.
\begin{algorithm}[htb]
\SetAlgoLined
\KwIn{Dataset $D$ of $n$ images, memory limit $M$, number of partition $k$, image neighborhood size $N$, $\nu^*$, $\tau$.}
Partition $D$ into random $k$ parts $D_1$, ..., $D_k$, each has roughly $\lfloor n/k \rfloor$ images. \\
Compute the maximum number of positive entries in the score matrices in each parts:
$K_1 \longleftarrow M/(N*\lfloor n/k \rfloor)$. \\
Compute the maximum number of positive entries in the score matrices in the whole dataset:
$K_2 \longleftarrow M/(n*N)$. \\
\For{$i = 1$ to $k$}{
  Compute score matrices for image pairs in $D_i$ with $K_1$ positive entries. \\
  Run proxy OSD on $D_i$ with $\nu = K_2$.\\
  Each image in $D_i$ has a new set of region proposals which are those retained by OSD.
}
Compute score matrices between pairs of images in $D$ with $K_2$ positive entries.\\
Run OSD on the whole dataset $D$ with $\nu=\nu^*$.
\caption{Large-scale object discovery algorithm.}
\label{algo:large_scale_OSD}
\end{algorithm}

\section{Experimental Results}

\subsection{Results with the Ensemble Method from~\cite{Vo2019UnsupOptim}}

Vo {\em et al.}~\cite{Vo2019UnsupOptim} use an ensemble method (EM) to combine several solutions before post processing to stabilize and improve the final performance of OSD. We investigate the influence of this procedure on the performance of OSD and rOSD with our proposals, and present the result in Tables~\ref{table:em1} and \ref{table:em2}. We use VGG16 features in these experiments. It can be seen that the effect of EM is mixed for the tested datasets. It generally harms the performance on VOC\_all and VOC12 and improves the performance on VOC\_6x2 while its effect on OD is unclear. We have therefore chosen to omit EM in the experiments of the main body of the paper.

\begin{table}[]
    \centering
    \caption{Influence of the ensemble method of Vo {\em et al.} on the colocalization performance of OSD and rOSD with our proposals}
    \label{table:em1}
    \resizebox{\textwidth}{!}{%
    \setlength{\tabcolsep}{1em}
    \begin{tabular}{cccccc}
        \toprule
        \multicolumn{2}{c}{Method} & OD & VOC\_6x2 & VOC\_all & VOC12 \\
        \midrule
        Ours (OSD) & w/o EM & \underline{89.0 $\pm$ 0.6} & 73.6 $\pm$ 0.6 & 44.7 $\pm$ 0.3 & \underline{49.0 $\pm$ 0.2} \\
        Ours (OSD) & w/ EM & 88.2 $\pm$ 0.2 & \textbf{75.3 $\pm$ 0.2} & 44.7 $\pm$ 0.1 & 48.7 $\pm$ 0.1 \\
        \midrule
        Ours (rOSD) & w/o EM & \underline{89.0 $\pm$ 0.5} & 73.3 $\pm$ 0.5 & \textbf{45.8 $\pm$ 0.3} & \textbf{49.7 $\pm$ 0.1} \\
        Ours (rOSD) & w/ EM & \textbf{89.2 $\pm$ 0.3} & \underline{74.5 $\pm$ 0.2} & \underline{45.5 $\pm$ 0.1} & \textbf{49.7 $\pm$ 0.2} \\
        \bottomrule
    \end{tabular}
    }
\end{table}

\begin{table}[]
    \centering
    \caption{Influence of the ensemble method of Vo {\em et al.} on the single-object discovery performance of OSD and rOSD with our proposals}
    \label{table:em2}
    \resizebox{\textwidth}{!}{%
    \setlength{\tabcolsep}{1em}
    \begin{tabular}{cccccc}
        \toprule
        \multicolumn{2}{c}{Method} & OD & VOC\_6x2 & VOC\_all & VOC12 \\
        \midrule
        Ours (OSD) & w/o EM & \underline{87.8 $\pm$ 0.4} & 69.2 $\pm$ 0.5 & \underline{48.7 $\pm$ 0.3} & 51.3 $\pm$ 0.2 \\
        Ours (OSD) & w/ EM & 87.5 $\pm$ 0.3 & 70.9 $\pm$ 0.3 & 48.6 $\pm$ 0.1 & 50.7 $\pm$ 0.1 \\
        \midrule
        Ours (rOSD) & w/o EM & 87.6 $\pm$ 0.3 & \underline{71.1 $\pm$ 0.8} & \textbf{49.2 $\pm$ 0.2} & \textbf{52.1 $\pm$ 0.1} \\
        Ours (rOSD) & w/ EM & \textbf{88.7 $\pm$ 0.3} & \textbf{71.9 $\pm$ 0.4} & \underline{48.7 $\pm$ 0.1} & \underline{52.0 $\pm$ 0.1} \\
        \bottomrule
    \end{tabular}
    }
\end{table}

\subsection{Full Results with both VGG16 and VGG19 Features}
We present in Tables~\ref{table:single_separate_supp},~\ref{table:single_mixed_supp} and~\ref{table:multiple_perf_all_supp} our full results in colocalization and object discovery with features from both VGG16 and VGG19. It can be seen that, with VGG16 features, rOSD still significantly outperforms OSD on the two large datasets and fares comparably to OSD on the smaller two. It is also noticeable that rOSD significantly outperforms Wei {\em et al.} in both colocalization and single-object discovery on all datasets when VGG16 features are used.

\begin{table*}[tb]
    \centering
    \caption{\small Single-object colocalization performance of our approach compared to the state of the art. Note that Wei {\em et al.}\,\protect{\cite{Wei2019ddtplus}} outperform our method on VOC\_all and VOC12 with VGG19 features in this case, but the situation is clearly reversed in the much more difficult single-object discovery setting, as demonstrated in Table~\ref{table:single_mixed_supp}}
    \label{table:single_separate_supp}
    \resizebox{\textwidth}{!}{%
    \setlength{\tabcolsep}{1em}
    \begin{tabular}{cccccc}
        \toprule
        Method & Features & OD & VOc\_6x2 & VOC\_all & VOC12 \\
        \midrule
        Cho {\em et al.}~\cite{CKSP15} & WHO & 84.2 & 67.6 & 37.6 & - \\
        Vo {\em et al.}~\cite{Vo2019UnsupOptim} & WHO & 87.1 $\pm$ 0.5 & 71.2 $\pm$ 0.6 & 39.5 $\pm$ 0.1 & - \\
        \midrule
        Li {\em et al.}~\cite{Li2016mimick} & VGG16 & - & - & 40.0 & 41.9 \\
        Wei {\em et al.}~\cite{Wei2019ddtplus} & VGG16 & 86.9 & 66.2 & 44.7 & 47.6 \\
        Ours (OSD) & VGG16 & 89.0 $\pm$ 0.6 & 73.6 $\pm$ 0.6 & 44.7 $\pm$ 0.3 & 49.0 $\pm$ 0.2 \\
        Ours (rOSD) & VGG16 & 89.0 $\pm$ 0.5 & 73.3 $\pm$ 0.5 & 45.8 $\pm$ 0.3 & \underline{49.7 $\pm$ 0.1} \\
        \midrule
        Li {\em et al.}~\cite{Li2016mimick} & VGG19 & - & - & 41.9 & 45.6 \\
        Wei {\em et al.}~\cite{Wei2019ddtplus} & VGG19 & 87.9 & 67.7 & \textbf{48.7} & \textbf{51.1} \\
        Ours (OSD) & VGG19 & \textbf{90.3 $\pm$ 0.3} & \underline{75.3 $\pm$ 0.7} & 45.6 $\pm$ 0.3 & 47.8 $\pm$ 0.2 \\
        Ours (rOSD) & VGG19 & \underline{90.2 $\pm$ 0.3} & \textbf{76.1 $\pm$ 0.7} & \underline{46.7 $\pm$ 0.2} & 49.2 $\pm$ 0.1 \\
        \bottomrule
    \end{tabular}
    }
\end{table*}

\begin{table*}[tb]
    \centering
    \caption{\small Single-object discovery performance in the mixed setting on the datasets with our proposals compared to the state of the art}
    \label{table:single_mixed_supp}
    \resizebox{\textwidth}{!}{%
    \setlength{\tabcolsep}{1em}
    \begin{tabular}{cccccc}
        \toprule
        Method  & Features & OD & VOC\_6x2 & VOC\_all & VOC12 \\ 
        \midrule
        Cho {\em et al.}~\cite{CKSP15} & WHO & 82.2 & 55.9 & 37.6 & -\\ 
        Vo {\em et al.}~\cite{Vo2019UnsupOptim} & WHO & 82.3 $\pm$ 0.3 & 62.5 $\pm$ 0.6 & 40.7 $\pm$ 0.2 & - \\
        \midrule
        Wei {\em et al.}~\cite{Wei2019ddtplus} & VGG16 & 73.5 & 66.2 & 41.9 & 45.0 \\
        Ours (OSD) & VGG16 & 87.8 $\pm$ 0.4 & 69.2 $\pm$ 0.5 & 48.7 $\pm$ 0.3 & 51.3 $\pm$ 0.2 \\
        Ours (rOSD) & VGG16 & 87.6 $\pm$ 0.3 & 71.1 $\pm$ 0.8 & \underline{49.2 $\pm$ 0.2} & \textbf{52.1 $\pm$ 0.1} \\
        \midrule
        Wei {\em et al.}~\cite{Wei2019ddtplus} & VGG19 & 75.0 & 54.0 & 43.4 & 46.3 \\
        Ours (OSD) & VGG19 & \underline{89.1 $\pm$ 0.4} & \underline{71.9 $\pm$ 0.7} & 47.9 $\pm$ 0.3 & 49.2 $\pm$ 0.2 \\
        Ours (rOSD) & VGG19 & \textbf{89.2 $\pm$ 0.4} & \textbf{72.5 $\pm$ 0.5} & \textbf{49.3 $\pm$ 0.2} & \underline{51.2 $\pm$ 0.2} \\
        \bottomrule
    \end{tabular}
    }
\end{table*}

\begin{table}[tb]
    \centering
    \caption{\small Multi-object colocalization and discovery performance of rOSD compared to competitors on VOC\_all and VOC12 datasets}
    \label{table:multiple_perf_all_supp}
    \resizebox{\textwidth}{!}{%
    \setlength{\tabcolsep}{1em}
    \begin{tabular}{cccccc}
        \toprule
        \multirow{2}{*}{Method} & \multirow{2}{*}{Features} & \multicolumn{2}{c}{Colocalization} & \multicolumn{2}{c}{Discovery} \\
         & & VOC\_all & VOC12 & VOC\_all & VOC12 \\
        \midrule
        Vo {\em et al.}~\cite{Vo2019UnsupOptim} & WHO & 40.7 $\pm$ 0.1 & - & 30.7 $\pm$ 0.1 & - \\
        \midrule
        Wei {\em et al.}~\cite{Wei2019ddtplus} & VGG16 & 38.3 & 40.4 & 25.8 & 28.2 \\
        Ours (OSD) & VGG16 & 45.9 $\pm$ 0.1 & 48.1 $\pm$ 0.0 & 34.9 $\pm$ 0.1 & 37.6 $\pm$ 0.0 \\
        Ours (rOSD) & VGG16 & \underline{48.5 $\pm$ 0.1} & \underline{50.7 $\pm$ 0.1} & \underline{37.2 $\pm$ 0.1} & \textbf{40.8 $\pm$ 0.1} \\
        \midrule
        Wei {\em et al.}~\cite{Wei2019ddtplus} & VGG19 & 43.3 & 45.5 & 28.1 & 30.3 \\
        Ours (OSD) & VGG19 & 46.8 $\pm$ 0.1 & 47.9 $\pm$ 0.0 & 34.8 $\pm$ 0.0 & 36.9 $\pm$ 0.0 \\
        Ours (rOSD) & VGG19 & \textbf{49.4 $\pm$ 0.1} & \textbf{51.5 $\pm$ 0.1} & \textbf{37.6 $\pm$ 0.1} & \underline{40.4 $\pm$ 0.1} \\
        \bottomrule
    \end{tabular}
    }
\end{table}

\subsection{Multi-Object Experiments}
\begin{figure*}
\centering
\begin{subfigure}{0.48\textwidth}
    \includegraphics[width=\linewidth]{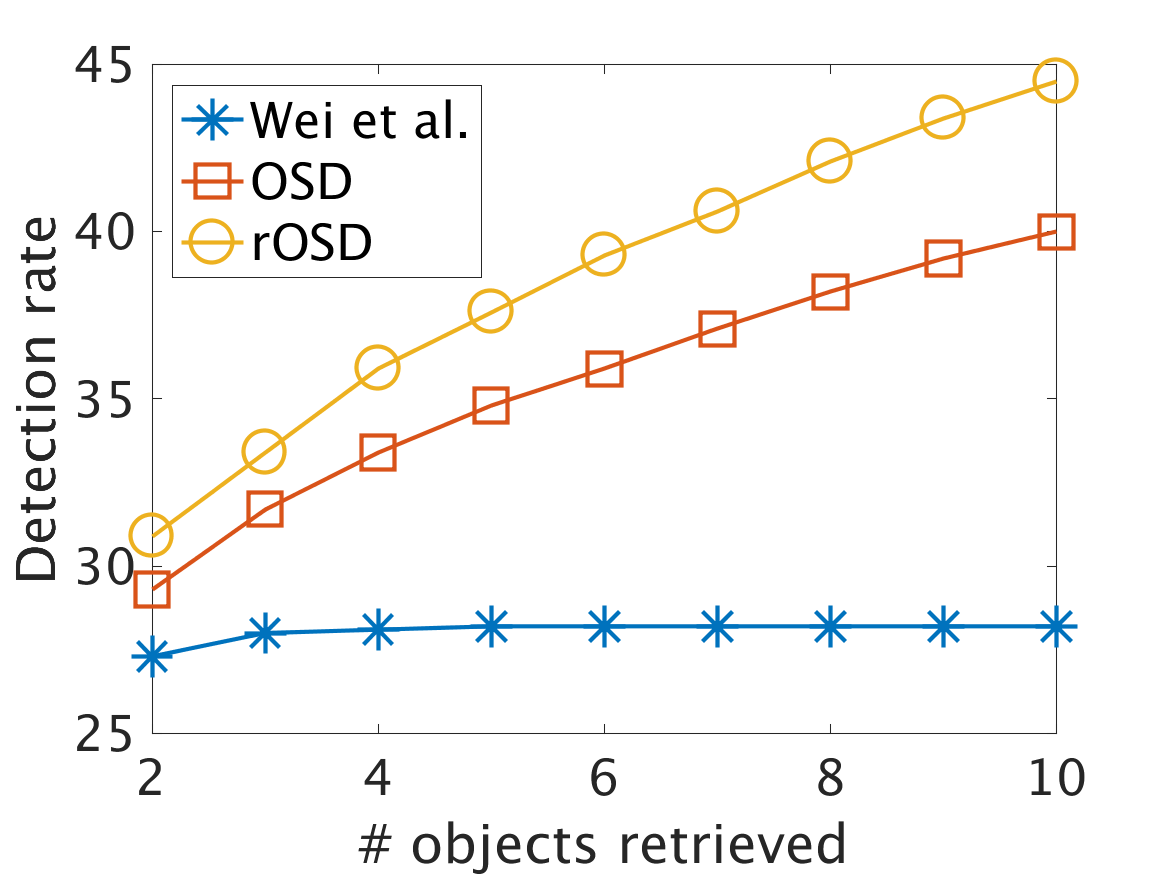}
    \caption{VOC\_all}
\end{subfigure}
\begin{subfigure}{0.48\textwidth}
    \includegraphics[width=\linewidth]{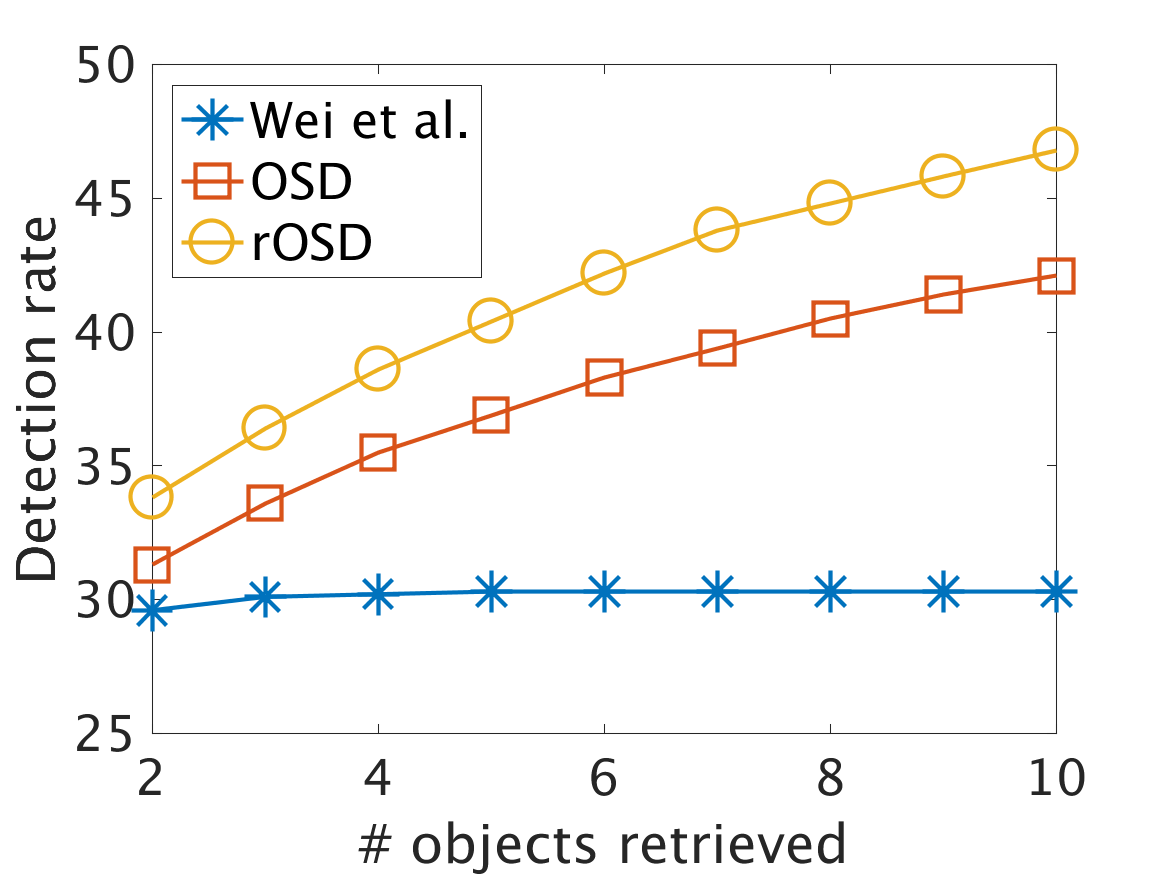}
    \caption{VOC12}
\end{subfigure}
\caption{\small Multi-object discovery performance of rOSD compared to OSD and~\cite{Wei2019ddtplus} when varying the maximum number of returned objects.}
\label{fig:multi_object_vary_numbox}
\end{figure*}

For a fair comparison to OSD and Wei {\em et al.}~\cite{Wei2019ddtplus} in multi-object discovery, we have fixed the number of objects retained in each image by all methods to $5$ in the paper. We have also modified the method of Wei {\em et al.} such that $5$ bounding boxes around the $5$ largest clusters of positive pixels in their \textit{indicator matrix} are returned as objects. 
For OSD and rOSD, we run the corresponding optimization then apply the following post processing on each image: all $\nu$ retained regions are ranked in descending order using the score proposed in \cite{Vo2019UnsupOptim} (Eq.\,12 in Sec.\,2.6 therein), which is solely based on their similarity to the retained regions in the image's neighbors; We then iteratively discard all proposals having an IoU score greater than some threshold with higher-ranked regions; Among remaining regions, we return the 5 highest ranked as retrieved objects. 
Since this procedure can eliminate all but a few regions if the regions highly overlap, we choose a large value of $\nu$ ($50$) and a large value of $IoU$ threshold ($0.7$) in our experiments to guarantee that we have \textit{exactly} 5 objects. 
%
%
This is, however, just a design choice and one can choose to retain fewer or more regions. We have conducted experiments with the number of retrieved objects varied in the interval $[2,10]$ and observed that rOSD always yields better performance than OSD and~\cite{Wei2019ddtplus} regardless of the number of objects retrieved (Fig.~\ref{fig:multi_object_vary_numbox}).

\begin{figure*}
\centering
\parbox[t][][t]{\linewidth}{
\includegraphics[width=\linewidth]{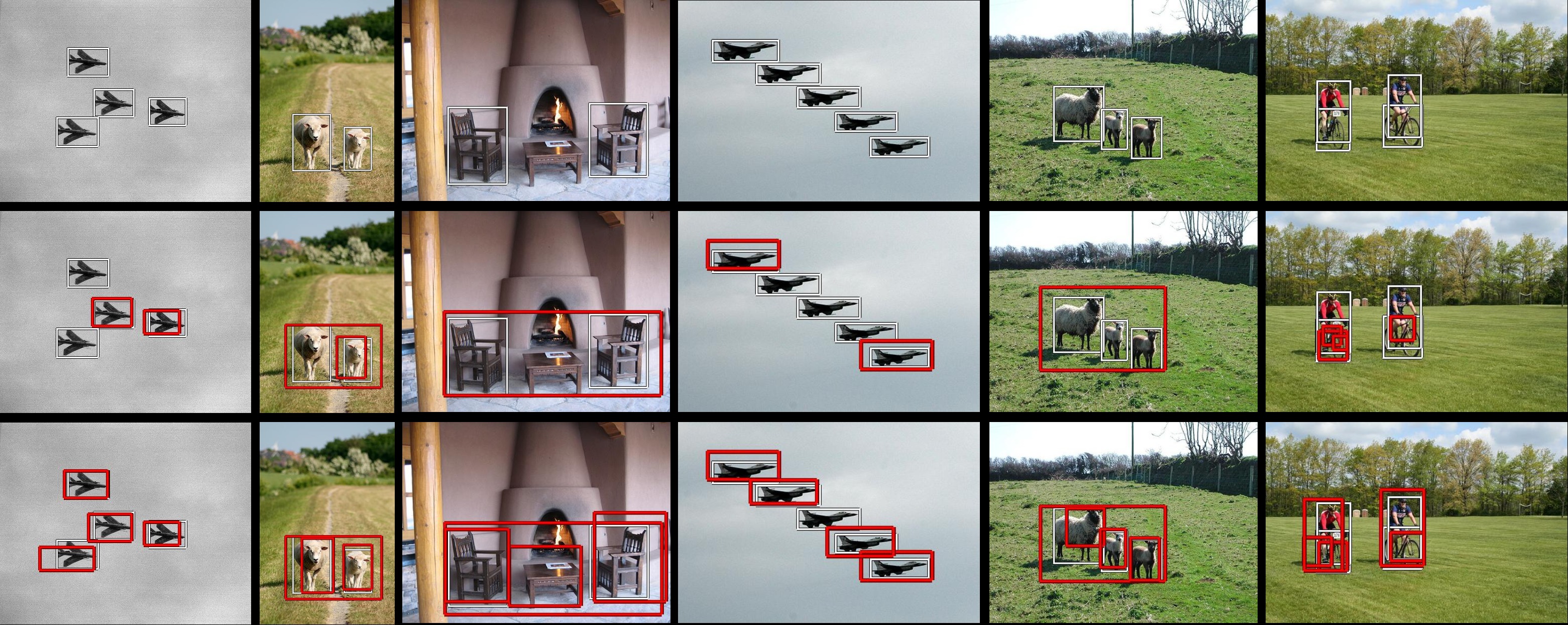}
}
\caption{\small Multi-object discovery results. In each column, from top to bottom: original image, image with predictions of OSD, image with predictions of rOSD. White boxes are ground truth objects and red ones are our predictions. There are \textit{at most} 5 predictions per image.}
\label{fig:multi_object}
\end{figure*}

Images may of course contain fewer than 5 objects. In such cases, OSD and rOSD usually return overlapping boxes around the actual objects (Fig. 3 in the paper). We can eliminate these overlapping boxes and obtain better qualitative results by using smaller $\nu$ and $IoU$ threshold. We have conducted preliminary experiments with $\nu=25$ in the optimization of OSD and rOSD and $IoU=0.3$ for suppression threshold 
in the post processing and show qualitative results in Fig.~\ref{fig:multi_object}. It can be seen that rOSD is now able to return bounding boxes around objects without many overlapping regions. It is also observed that rOSD fares much better than OSD in localizing multiple objects. We also compare the quantitative performance of rOSD, OSD and~\cite{Wei2019ddtplus} in Table~\ref{table:multiple_perf_all_supp_new}. For~\cite{Wei2019ddtplus}, we take as before the bounding boxes around the largest clusters of pixels in the \textit{indicator matrix} of each image. The number of clusters in this case is chosen to be the number of objects returned by rOSD in the same image. The results show that rOSD again yields by far the best performance. It is also noticeable that while using smaller values of $\nu$ and the $IoU$ threshold slightly deteriorates the performance of rOSD, it makes the performance of OSD drop significantly (compare Tables~\ref{table:multiple_perf_all_supp} and~\ref{table:multiple_perf_all_supp_new}). This is due to the fact that OSD returns many highly overlapping regions and most of them are eliminated by our procedure. 
On the other hand, rOSD returns more diverse regions and consequently more regions are retained. 
In practice, we observe that OSD returns on average 1.47 (respectively 1.52) regions while rOSD returns 3.62 (respectively 3.63) on VOC\_all (respectively VOC12). Note, however, that rOSD still outperforms OSD and~\cite{Wei2019ddtplus} even when the latter are allowed to retain exactly 5 regions.

\begin{table}[tb]
    \centering
    \caption{\small Multi-object colocalization and discovery performance of rOSD compared to competitors on VOC\_all and VOC12 datasets when using smaller values of $\nu$ ($25$) and $IoU$ ($0.3$) threshold}
    \label{table:multiple_perf_all_supp_new}
    \resizebox{\textwidth}{!}{%
    \setlength{\tabcolsep}{1em}
    \begin{tabular}{cccccc}
        \toprule
        \multirow{2}{*}{Method} & \multirow{2}{*}{Features} & \multicolumn{2}{c}{Colocalization} & \multicolumn{2}{c}{Discovery} \\
         & & VOC\_all & VOC12 & VOC\_all & VOC12 \\
        \midrule
        Wei {\em et al.}~\cite{Wei2019ddtplus} & VGG19 & \underline{43.1} & \underline{45.3} & 27.8 & 30.0 \\
        Ours (OSD) & VGG19 & 39.6 $\pm$ 0.1 & 41.6 $\pm$ 0.1 & \underline{29.0 $\pm$ 0.1} & \underline{31.3 $\pm$ 0.1} \\
        Ours (rOSD) & VGG19 & \textbf{47.3 $\pm$ 0.1} & \textbf{49.3 $\pm$ 0.1} & \textbf{36.7 $\pm$ 0.1} & \textbf{39.2 $\pm$ 0.1} \\
        \bottomrule
    \end{tabular}
    }
    \vspace{-10pt}
\end{table}

\subsection{Evaluating the Graph Computed by OSD}

Following~\cite{CKSP15}, we evaluate the local graph structure obtained by rOSD using the CorRet measure, defined as the average percentage of returned image neighbors that belong to the same (ground-truth) class as the image itself. As a baseline, we consider the local graph induced by the sets of nearest neighbors $N(i)$ computed from the fully connected layer \textit{fc6} of the CNN that are used in the same experiment. Table~\ref{table:corret} shows the CorRet of local graphs obtained when running rOSD (OSD) on VOC\_all and VOC12 and large-scale rOSD (OSD) on COCO\_20k in the mixed setting. It can be seen that the local image graphs returned by our methods have higher CorRet than the baseline.
\begin{table}[tb]
    \centering
    \caption{\small Quality of the returned local image graph as measured by CorRet}
    \label{table:corret}
    \setlength{\tabcolsep}{1em}
    \begin{tabular}{cccc}
        \toprule
        Dataset & VOC\_all & VOC12 & COCO\_20k  \\
        \midrule
        Baseline & 50.7 & 56.4 & 36.8 \\
        Ours (OSD) & \textbf{60.1 $\pm$ 0.1} & \textbf{63.2 $\pm$ 0.0} & \textbf{39.8 $\pm$ 0.0} \\
        Ours (rOSD) & \underline{59.8 $\pm$ 0.1} & \underline{63.0 $\pm$ 0.0} & \underline{39.4 $\pm$ 0.0} \\
        \bottomrule
    \end{tabular}
\end{table}

\subsection{Results on Images of ImageNet Classes not in the Training Set of the Feature Extractors}
Though trained for classifying 1000 object classes of ImageNet, features from convolutional layers of VGGs have shown to be generic: They have been used for various tasks, including unsupervised object discovery. Li {\em et al.}~\cite{Li2016mimick} and Wei {\em et al.}~\cite{Wei2019ddtplus} have shown that CNN features generalize well beyond the classes in ILSVRC2012 by testing on 6 held-out classes on ImageNet (\textit{chipmunk}, \textit{racoon}, \textit{rhinoceros}, \textit{rake}, \textit{stoat} and \textit{wheelchair}). We have also tested our method on these classes. Since ImageNet has been under maintenance, we could not download all the official images in the six classes. For preliminary experiments, we have instead downloaded the images using their public URLs (provided on the ImageNet website), eliminated corrupted images, randomly chosen up to 200 images per class and run our experiments on these images. We have compared rOSD, OSD,~\cite{Li2016mimick} and~\cite{Wei2019ddtplus} in this setting (Table~\ref{table:imagenet}). Although rOSD performs significantly better than~\cite{Li2016mimick} in colocalization tasks, it is as before significantly outperformed by~\cite{Wei2019ddtplus} there.  In object discovery, rOSD performs slightly better than~\cite{Wei2019ddtplus} for VGG16 features, but significantly worse for VGG19 features. Understanding this discrepancy observed in preliminary experiments is part of our plans for future work.

\begin{table}[tb]
    \centering
    \caption{\small Colocalization and single-object discovery performance of rOSD compared to OSD, Li {\em et al.}~\cite{Li2016mimick} and  Wei {\em et al.}~\cite{Wei2019ddtplus} on 6 \textit{held-out} ImageNet classes}
    \label{table:imagenet}
    \resizebox{0.75\textwidth}{!}{%
    \setlength{\tabcolsep}{1em}
    \begin{tabular}{cccc}
        \toprule
        Method & Features & Colocalization & Discovery \\
        \midrule
        Li {\em et al.}\footnote{Numbers for~\cite{Li2016mimick} are taken from~\cite{Wei2019ddtplus}.}~\cite{Li2016mimick} & VGG16 & 48.3 & - \\
        Wei {\em et al.}~\cite{Wei2019ddtplus} & VGG16 & \underline{74.3} & 61.2 \\
        Ours (OSD) & VGG16 & 61.5 $\pm$ 0.3 & 60.3 $\pm$ 0.3 \\ 
        Ours (rOSD) & VGG16 & 63.0 $\pm$ 0.7 & \underline{61.6 $\pm$ 0.4} \\
        \midrule
        Li {\em et al.}~\cite{Li2016mimick} & VGG19 & 51.6 & - \\
        Wei {\em et al.}~\cite{Wei2019ddtplus} & VGG19 & \textbf{74.8} & \textbf{63.2} \\
        Ours (OSD) & VGG19 & 61.3 $\pm$ 0.5 & 59.2 $\pm$ 0.7 \\ 
        Ours (rOSD) & VGG19 & 63.7 $\pm$ 0.3 & 59.4 $\pm$ 0.5 \\
        \bottomrule
    \end{tabular}
    }
    \vspace{-10pt}
\end{table}

\section{More Visualizations}
\subsection{Overlapping Regions Returned by OSD and rOSD}
\begin{figure}[htb]
    \centering
    \parbox[t][][t]{\linewidth}{
    \vspace{-0.7cm}
    \includegraphics[width=\linewidth]{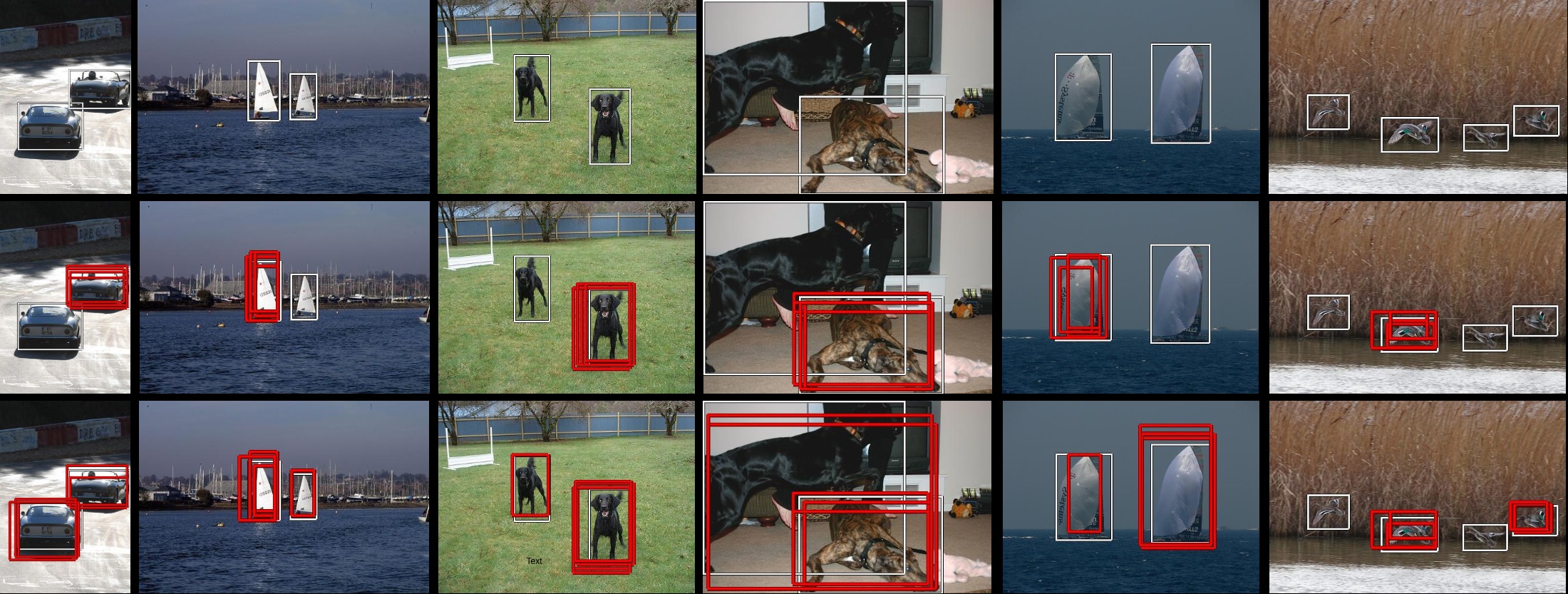}
    }
	\caption{\small Regions returned by OSD and rOSD. In each column, from top to bottom: original image, image with regions returned by OSD, image with regions returned by rOSD.}
	\label{fig:overlap}
\end{figure}

The most important advantage of rOSD over OSD is that the former returns more diverse regions than the former does. We visualize the regions returned by OSD and rOSD in colocalization experiments with $\nu=5$ in Fig.~\ref{fig:overlap}. 

\subsection{Persistence}
\begin{figure}[htb]
    \centering
    \vspace{-10pt}
        \includegraphics[width=0.8\linewidth]{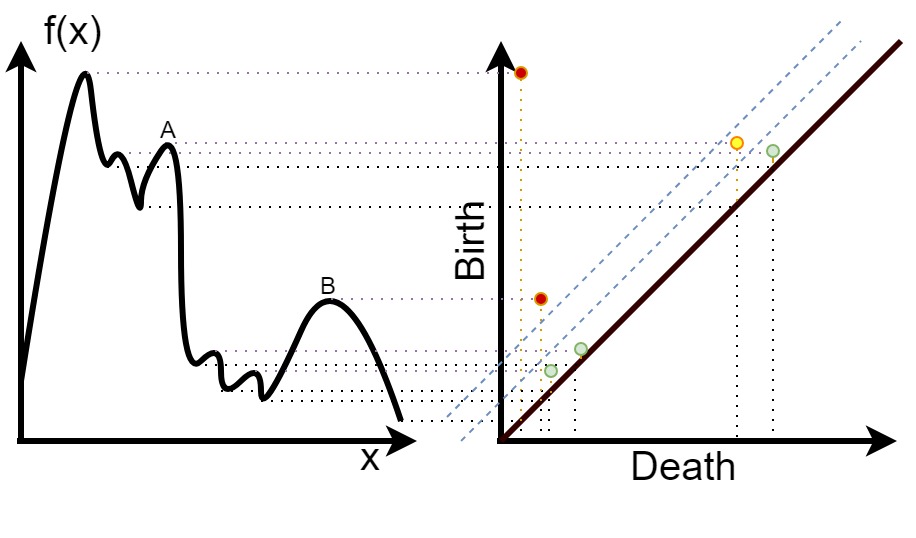}
	 \vspace{-5mm}
        \caption{\small An illustration of persistence in the 1D case. Left: A 1D function. Right: Its persistence diagram. Points above the diagonal correspond to its local maxima and the vertical distance from these points to the diagonal is their persistence. Local maxima with higher persistence are more robust: B is more robust than A although $f(A) > f(B)$. Given a chosen persistence threshold (shown by dash lines in blue), points with persistence higher than some threshold are selected as robust local maxima. The black horizontal dotted lines show birth and death time of the local maxima of $f$.}
	\vspace{-5mm}
	\label{fig:persistence}
\end{figure}

We use persistence~\cite{Chazal2013persistence,Edelsbrunner2009introtopo,Edelsbrunner2002topo,Oudot2015persistence,Zomorodian2005compute} to find robust local maxima of the global saliency map $s_g$ in our work. Considering $s_g$ as a 2D image and each location in it as a pixel, we associate with each pixel a cluster (the 4-neighborhood connected component of pixels that contains it), together with both a ``birth'' (its own saliency) and ``death time'' (the highest value for which one of the pixels in its cluster also belongs to the cluster of a pixel with higher saliency, or, if no such location exists, the lowest saliency value in the map). The persistence of a pixel is defined as the difference between its birth and death times. Figure~\ref{fig:persistence} illustrates persistence for the 1D case.

\end{document}